\documentclass[sigconf, screen]{acmart}

\usepackage{amssymb}
\usepackage{makecell}
\usepackage{multirow}
\usepackage{cuted} 
\usepackage{algorithm}
\usepackage{hyperref}
\usepackage{siunitx}
\usepackage[most]{tcolorbox}
\tcbset{
  promptbox/.style={
    colback=gray!5,
    colframe=gray!60,
    boxrule=0.4pt,
    arc=2pt,
    left=6pt, right=6pt, top=4pt, bottom=4pt,
    enhanced jigsaw,
  }
}
\usepackage{pifont}
\AtBeginDocument{%
  }

\setcopyright{acmlicensed}

\copyrightyear{2026}
\acmYear{2026}
\setcopyright{cc}
\setcctype{by-nc-nd}
\acmConference[MM '26]{Proceedings of the 34th ACM International Conference on Multimedia}{November 10--14, 2026}{Rio de Janeiro, Brazil}
\acmBooktitle{Proceedings of the 34th ACM International Conference on Multimedia (MM '26), November 10--14, 2026, Rio de Janeiro, Brazil}
\acmDOI{10.1145/3767308.3836218}
\acmISBN{979-8-4007-2213-4/2026/11}
\begin{document}
\title{TIQA: Human-Aligned Perceptual Text Quality Assessment in Generated Images}

\author{Kirill Koltsov}
\email{kirill.koltsov@graphics.cs.msu.ru}
\affiliation{%
  \institution{Lomonosov Moscow State University}
  \city{Moscow}
  \country{Russia}
}

\author{Aleksandr Gushchin}
\email{alexander.gushchin@graphics.cs.msu.ru}
\orcid{0002-4055-7394}
\affiliation{%
  \institution{Trusted AI Research Center RAS}
  \city{Moscow}
  \country{Russia}
}

\author{Dmitriy Vatolin}
\email{dmitriy@graphics.cs.msu.ru}
\affiliation{%
  \institution{MSU Institute for Artificial Intelligence}
  \city{Moscow}
  \country{Russia}
}

\author{Anastasia Antsiferova}
\email{aantsiferova@graphics.cs.msu.ru}
\affiliation{%
  \institution{Trusted AI Research Center RAS}
  \city{Moscow}
  \country{Russia}}

\begin{abstract}
    Recent text-to-image models have improved global realism, but text rendering remains a persistent failure mode: images may look convincing overall, yet local typography often contains malformed glyphs, broken strokes, irregular spacing, and other artifacts that humans heavily penalize. We formulate Text-in-Image Quality Assessment (TIQA), a no-reference task that estimates a human-aligned perceptual quality score for detected text regions independently of semantic correctness. We also introduce two datasets. TIQA-Crops contains 120k text crops from 36k AI-generated images from 12 generators, with 10k mean-opinion-score (MOS) labels and 110k proxy labels for pretraining. TIQA-Images contains 1,500 text-heavy images from 10 recent generators, including proprietary systems, with paired overall-quality and text-quality subjective scores. We also propose an AI-generated No-reference Text-in-Image Quality Assessment (ANTIQA) model, a lightweight predictor with text-specific inductive biases. Across crop-level and image-level evaluations, ANTIQA achieves the best alignment with human judgments, reaching PLCC/SROCC of 0.942/0.935 on TIQA-Crops and 0.842/0.837 for text-quality MOS on unseen generators in TIQA-Images. These results establish perceptual text quality as a distinct evaluation target for modern text-to-image generation. The code and dataset are available at \url{https://github.com/koltsov-cmc/antiqa}.
\end{abstract}






\maketitle

\section{Introduction}

\begin{figure}[t!]
    \centering
    \includegraphics[width=\linewidth]{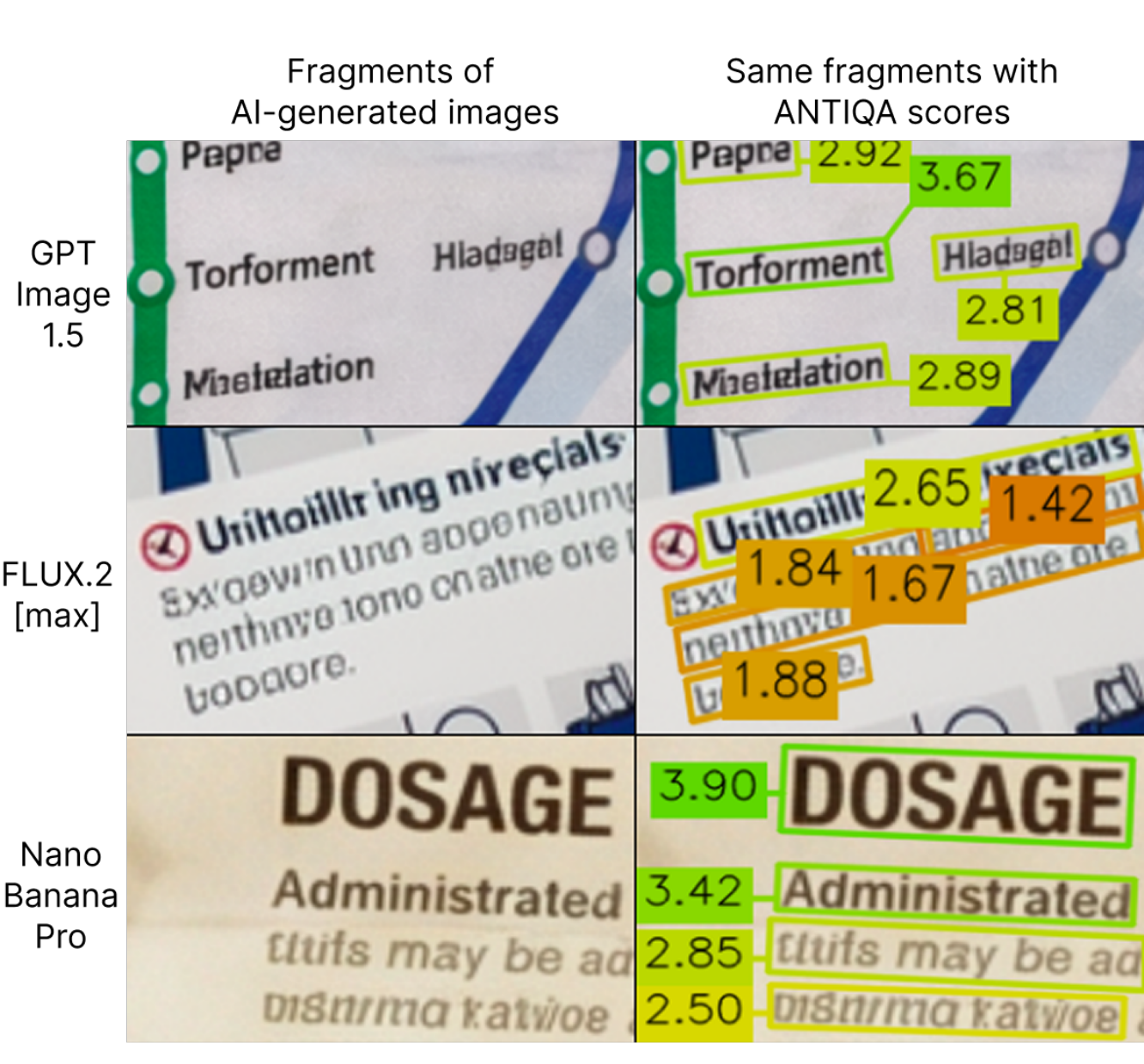}
    \caption{ANTIQA in action. Left: text-containing fragments from images generated by SOTA generators. Right: ANTIQA's output — a Perceptual Quality score predicted for each crop.}
    \label{fig:crop_examples}
    \Description{TODO}
\end{figure}

The rapid development of generative AI has made AI-generated images widely accessible, with text-to-image (T2I) systems becoming simultaneously faster, cheaper, and higher quality.
Recent models have improved substantially on prompt semantics and global realism, as reflected in modern benchmarks \cite{lmarena, li2024aigiqa, zhang2025q}.
Yet \emph{text rendering} remains a persistent failure mode: generated images often exhibit malformed glyphs, broken strokes, inconsistent thickness, and unstable kerning or baselines (Figure~\ref{fig:crop_examples}).
These errors are highly salient in text-heavy outputs (posters, UI mockups, pseudo-documents), yet current evaluations lack a dedicated way to measure the \emph{perceptual fidelity} of rendered text.

Existing approaches evaluate text in images either through \\ recognition-centric pipelines (e.g., OCR against ground-truth text) or by using a large vision--language model (VLM) as a general-purpose judge~\cite{textinvision,bosheah2025challenges}.
Both are useful, but neither reliably captures perceptual text quality.
OCR-based scores primarily reflect semantic correctness and require ground truth; they can under-penalize appearance defects that humans judge harshly (e.g., stroke breaks, irregular thickness, kerning/baseline instability) even when the string remains decodable.
VLMs can, in principle, reason about such artifacts across languages and styles, but their use as a benchmark faces well-known obstacles: (i) the output results from a prompting/decoding procedure and is sensitive to prompt wording, sampling, and preprocessing, making standardization difficult \cite{gu2024survey, zhu2024adaptive}; (ii) closed, frequently updated APIs may introduce version drift, so benchmark outcomes can vary over time without changes to the evaluated method; and (iii) even in adjacent perceptual-quality tasks, specialized quality models can outperform GPT-4V-style judges despite detailed instructions \cite{depictqa_v1,deqa_score}.
Together, these limitations motivate a dedicated model that directly targets perceptual text artifacts.

We address this gap by introducing \emph{Text-in-Image Quality Assessment} (TIQA): predicting a scalar score for a detected text region that matches human judgments of rendered-text fidelity, \emph{independent of semantic correctness}. We exclude semantic correctness by design, since our goal is a no-reference perceptual metric for rendered-text appearance; correctness is a complementary axis better measured by OCR or VLM-based recognition.
The main contributions of this work are as follows:

\begin{itemize}
    \item \textbf{New task formulation (TIQA).} Given a detected text region in an AI-generated image, the goal is to predict a single perceptual quality score aligned with human judgments of rendering artifacts (e.g., malformed glyphs, broken strokes, character hallucinations), \emph{independent of the semantic correctness of the text}. 
    
    \item \textbf{Two datasets for benchmarking and training.}
    \textbf{TIQA-Crops} contains 120k OCR-detected text crops from 12 T2I models: 10k with MOS labels and 110k used for \\ proxy-supervised pretraining via OCR confidence.
    \textbf{TIQA-Images} provides 1,500 full-frame, text-heavy images from 10 recent generators (e.g., GPT Image 1.5, Nano Banana Pro), each paired with a text-only view and dual MOS annotations (overall and text-only). Public links will be provided upon acceptance.
    
    \item \textbf{Method (ANTIQA).} We develop \textbf{ANTIQA}, a specialized TIQA model that outperforms strong baselines — \\ OCR-derived confidence, generic IQA metrics, and VLM-based judges — under both in-distribution and cross-generator evaluations, improving correlation over the second-best \\ method across all settings (e.g., $+0.08$ PLCC for the latest T2I models). 

    \item \textbf{Analysis and applications.} We characterize text rendering failures across modern T2I systems (including proprietary models) and quantify remaining gaps; we also show that TIQA scores enable effective filtering/best-of-$K$ selection and are useful on text-heavy images in downstream vision tasks, including AI-image detection.
\end{itemize}

\begin{figure*}[h!]
    \centering
    \includegraphics[width=\linewidth]{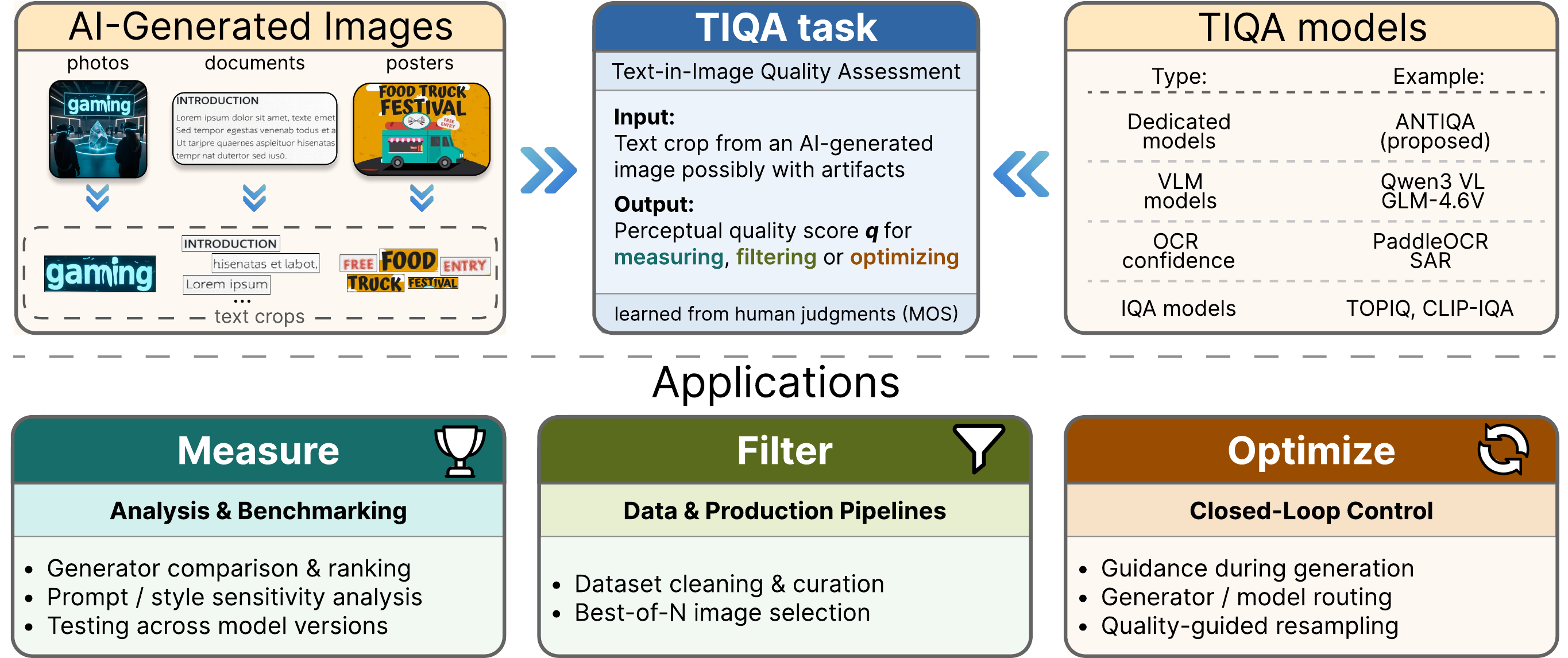}
    \caption{Overview of Text-in-Image Quality Assessment (TIQA). Left: AI-generated images contain multiple text regions that are detected and cropped. Middle: a TIQA model predicts a scalar text-quality score for each crop, trained on mean opinion scores (MOS). Right: representative model families used as baselines (VLM judges, OCR confidence, generic IQA) and the proposed specialized TIQA model. Bottom: example applications of TIQA for measuring generator quality, filtering candidates in production pipelines (best-of-K), and optimizing generation via reranking or closed-loop control.} 
    \label{fig:scheme}
    \Description{TODO}
\end{figure*}

\section{Related Work}
We study no-reference, crop-level prediction of perceptual text rendering quality in AI-generated images, targeting typographic artifacts (glyph topology/shape, stroke continuity/thickness, etc.) rather than semantic string correctness.

\textbf{Improving text rendering in generative models.}
Rendered text can be improved via text-aware conditioning, layout/glyph guidance, and post-editing/inpainting pipelines \cite{textgen_placeholder,text_guidance_placeholder,postedit_placeholder}, all of which benefit from reliable local feedback to train, guide, or rerank generations. TIQA provides this missing signal: a MOS-aligned, region-level score for typographic appearance, independent of semantic correctness, complementing OCR-based metrics and prompted-judge approaches.

\textbf{Image quality evaluation and AIGC evaluation.}
Image generators are commonly evaluated with distributional realism metrics (IS \cite{salimans2016improved}, FID \cite{heusel2017gans}) and, when references exist, full-reference fidelity (PSNR/SSIM, LPIPS \cite{zhang2018unreasonable}). In the no-reference setting, generic IQA spans blind-feature models (BRISQUE \cite{mittal2012no}), learned MOS predictors (NIMA \cite{talebi2018nima}), and transformer-based methods (TOPIQ \cite{topiq}). Preference/reward/judge scores (e.g., \cite{image_reward,pickscore}) assess semantic match, aesthetics, or overall quality, but are not designed to isolate fine-grained text artifacts that can dominate human judgments in text-heavy images while leaving global realism largely unchanged.

\textbf{Evaluating text in AI-generated images: correctness vs. appearance.}
Text evaluation for T2I outputs is often recognition-centric: OCR outputs are compared to prompts or ground truth (e.g., \cite{textinvision,zhang2025astrict}) using CER/Levenshtein metrics. Effective for decodability and semantic match, such measures can under-penalize perceptual defects (broken strokes, malformed glyph topology, unstable kerning/baselines) that humans rate poorly even when text is readable. VLM/LLM-based judging (e.g., \cite{bosheah2025challenges,sampaio2024typescore}) is more comprehensive, but the score is the outcome of a procedure (prompting, decoding, preprocessing, cropping) rather than a standardized metric, and can be sensitive to prompt wording and model/version drift, consistent with broader ``LLM-as-a-judge'' findings \cite{gu2024survey,zhu2024adaptive,depictqa_v1,deqa_score}. These issues are amplified in region-level text crops, where small preprocessing differences can alter perceived artifacts.

\textbf{Downstream use and adjacent text-centric IQA.}
Learned scorers increasingly curate data \cite{schuhmann2022laion5b}, rank best-of-$K$ samples \cite{pickscore,image_reward}, and provide reward signals for refinement \cite{lee2023aligning,eyring2024reno,image_reward}. Yet these workflows typically rely on generic IQA, prompt-alignment scorers, or correctness proxies (OCR confidence/string match), poorly matched to typographic failure modes. Adjacent document/screen-content IQA and text legibility works \cite{diqa_placeholder,sciqa_placeholder,legibility_placeholder} address physical degradations (blur/compression) but not characteristic generative failures (hallucinated strokes, glyph topology, style-inconsistent character formation), nor a MOS-aligned signal calibrated to typographic plausibility. TIQA complements these directions by focusing on generative artifacts in detected text regions.

\section{Text-in-Image Quality Assessment (TIQA) for AI images}

Rendered text quality has at least two distinct dimensions: semantic correctness and perceptual rendering quality. This work focuses on the latter and introduces \emph{Text-in-Image Quality Assessment} (TIQA), a task in which a detected text crop is assigned a scalar score reflecting the perceptual quality of the rendered text, as judged by humans. TIQA targets visual properties such as glyph formation, stroke continuity, spacing, and typographic coherence, rather than whether the text is linguistically correct. By separating visual rendering attributes from language-level correctness, TIQA formalizes a complementary task that addresses aspects of rendered text not explicitly targeted by OCR-based measures and general VLM-based judges. Figure~\ref{fig:scheme} summarizes the TIQA task, representative model families, and downstream applications in measuring, filtering, and optimizing text-in-image generation. While both semantic correctness and perceptual fidelity matter for evaluating rendered text, prior work has focused mainly on the former. TIQA addresses this underexplored perceptual dimension and, together with already established semantic evaluation methods, supports a more complete assessment of rendered text.

\subsection{Task definition}

We propose text-in-image quality assessment (TIQA), a specialization of no-reference image quality assessment (IQA) for rendered text. Classical IQA aims to predict how an image appears to humans under common distortions (e.g., blur, noise, compression). In contrast, TIQA focuses on generator-induced text artifacts that corrupt the appearance of text in AI-generated images. Crucially, TIQA is \emph{independent of semantic correctness}: it evaluates how the text is rendered, not what the text says. Thus, a semantically correct string with perceptual rendering artifacts (e.g., malformed glyphs, broken strokes) must receive a lower TIQA score than a visually clean but misspelled string. TIQA is defined at the level of perceptual rendered-text quality; in this paper, we evaluate that problem in the Latin-script setting, leaving broader script coverage to future work.

Formally, given a text crop $x \in \mathcal{X} \subseteq \mathbb{R}^{H \times W \times 3}$, TIQA model predicts a scalar score $f(x) \in \mathbb{R}$ that correlates with the mean opinion score (MOS) $s(x)$ of rendered-text quality. We learn $f$ by minimizing the expected loss function $\ell$
\begin{equation}
    \min_{f} \; \mathbb{E}_{x \sim \mathcal{X}}\big[\ell(f(x), s(x))\big],    
\end{equation}
where $s(x)$ reflects the severity of AI-specific text artifacts rather than classical camera/codec degradations. For text crops, such artifacts primarily violate: (i) glyph integrity (character topology and stroke continuity), (ii) typographic regularity (spacing, alignment, baselines, consistent font style), and (iii) scene binding (physically consistent compositing on surfaces, perspective, and illumination). 
Examples of text crops with artifacts are shown in Figure~\ref{fig:crop_examples}. Additional examples are provided in Appendix~\ref{app:data}. We propose evaluating the TIQA model performance using standard correlation metrics on MOS-annotated datasets: Pearson’s Linear Correlation Coefficient (PLCC) and Spearman’s Rank Order Correlation Coefficient (SROCC).

\subsection{Downstream Tasks}\label{sec:downstream}

Beyond benchmarking, TIQA models provide a control signal that can be used throughout text-heavy generation pipelines: for data curation, as guidance during training or sampling, and at inference time for filtering and quality-aware routing of OCR/VLM reasoning when outcomes depend on rendered text. We highlight five representative use cases:

\begin{enumerate}
    \item \textbf{Reranking and filtering:} Rank multiple candidates per prompt by predicted text quality, or apply accept/reject thresholds. If all candidates fall below a threshold, resample up to a fixed budget and return the best, reducing visually corrupted text without changing the base generator.
    \item \textbf{Quality-aware routing for OCR/VLM reasoning:} When synthetic artifacts (malformed glyphs, inconsistent strokes, broken spacing, hallucinated characters) cause OCR/VLM failures, TIQA can (i) gate OCR outputs (accept vs.\ abstain), (ii) trigger re-generation/re-rendering (e.g., new seed / typography / layout), and (iii) pre-check text-dependent VQA to abstain or fall back to OCR-assisted reasoning when text is unlikely to be reliable.
    \item \textbf{AI-image detection as a complementary cue}: TIQA scores can be fused with general real-vs-AI detectors to provide an additional text-specific signal. In images containing text, perceptual text artifacts captured by TIQA may complement generic forensic cues and improve real-vs-AI classification.
    \item \textbf{Guidance for T2I models:} Use TIQA as a reward for selection among samples, or as an auxiliary objective during sampling/training to improve rendered text while keeping prompt semantics fixed.
    \item \textbf{Data curation for training:} Filter or stratify text-containing samples for OCR/VLM training to remove severe degradations, control difficulty (curricula or balanced sampling), and reduce noisy supervision from incoherent text.
\end{enumerate}

We evaluate reranking in Section~\ref{sec:reranking}. Appendix~B further demonstrates that a specialized TIQA model provides a complementary cue for AI-image detection and is predictive of failures in OCR and VLM-based vision tasks.

\section{Datasets for TIQA}
For training and analysis, we propose two datasets: \textbf{TIQA-Crops} and \textbf{TIQA-Images}. TIQA-Crops contains 120,000 cropped text regions extracted from 36,000 AI-generated images produced by a diverse set of 12 T2I models. We annotate 110,000 crops with OCR confidence, used only for pretraining; the remaining 10,000 are annotated with MOS of perceptual text quality, enabling supervised training and in-domain evaluation.
TIQA-Images is designed to analyze TIQA behavior and characterize modern T2I models on text-heavy prompts. Unlike the localized crops of TIQA-Crops, it consists of full-frame images (the entire generation, without cropping): 1,500 images from 10 T2I models, including proprietary systems (e.g., Nano Banana Pro, GPT Image 1.5), each annotated with two image-level MOSes, overall quality and text-only quality. For both datasets, each element is annotated with at least 50 ratings; details are provided below and in Appendix~\ref{app:human}.
\subsection{TIQA-Crops dataset: training and in-domain evaluation}
\textbf{Data Collection.}
We use a prompt dataset from TextInVision \cite{textinvision} to construct a large-scale, diverse dataset of AI-generated text artifacts in images. It provides 50,000+ methodically designed text-in-image prompts spanning simple, complex, and real-world scenarios (e.g., ads and educational materials), with prompt complexity and text attributes independently varied. Text strings are grouped into single words, phrases, and long multi-sentence text, with controlled difficulty (Oxford 5,000 CEFR A1–C1) and stress cases such as gibberish, misspellings, numbers, and special characters. 
We sampled 3,000 prompts and generated 36,000 AI images using 12 T2I models, from which we extracted 120,000 text regions using the PP-OCRv5 \cite{paddle} text detection model. We selected PP-OCRv5 based on an in-lab markup showing that, in 98\% of annotated images, all text-containing areas were detected correctly. The resulting crops include both clean text and diverse generation-induced artifacts. We also tested EasyOCR~\cite{easyocr} and RapidOCR~\cite{rapidocr}, but their performance was substantially lower (94\% and 89\%, respectively). More details about the prompts, in-lab markup, the list of T2I models, and examples of the final crops are provided in Appendix~\ref{app:data}.

\textbf{Human Annotation.}
To collect subjective quality scores, we used the Yandex.Tasks platform~\cite{yandex_tasks}. We designed a 0--5 text-quality scale, where 0 indicates no text (such crops were filtered out) and 5 indicates ideal quality. Participants were instructed to evaluate \emph{visual artifacts in the rendered text}, while \textit{ignoring meaning or spelling as much as possible}, since these aspects can be evaluated by OCR and VLM models. To guide raters, we provided detailed instructions, descriptions, and visual examples for each score from 0 to 5. To be eligible, subjects had to pass a 10-question exam with evenly distributed ground-truth scores, answering at least 8 correctly, and we further filtered low-quality responses with verification questions. In total, for 10,000 text crops, we collected 500,000+ scores from $\sim$4,500 unique participants. For the full instructions, statistics, inter-rater agreement and other details, see Appendix~\ref{app:human}.

\subsection{TIQA-Images: Text-Heavy Images from Modern T2I Models}
\label{sec:TIQA_images}

To complement our crop-level training data, we introduce \textbf{TIQA-Images}, a text-heavy benchmark of full AI-generated images. TIQA-Images is designed to analyze (i) how well TIQA models generalize to \emph{unseen} generators and prompts, and (ii) how overall image quality relates to the perceptual quality of rendered text. To help disentangle text artifacts from surrounding visual content, we additionally construct a paired text-only view for each image (described below).

\begin{figure*}[t!]
    \centering
    \includegraphics[width=\linewidth]{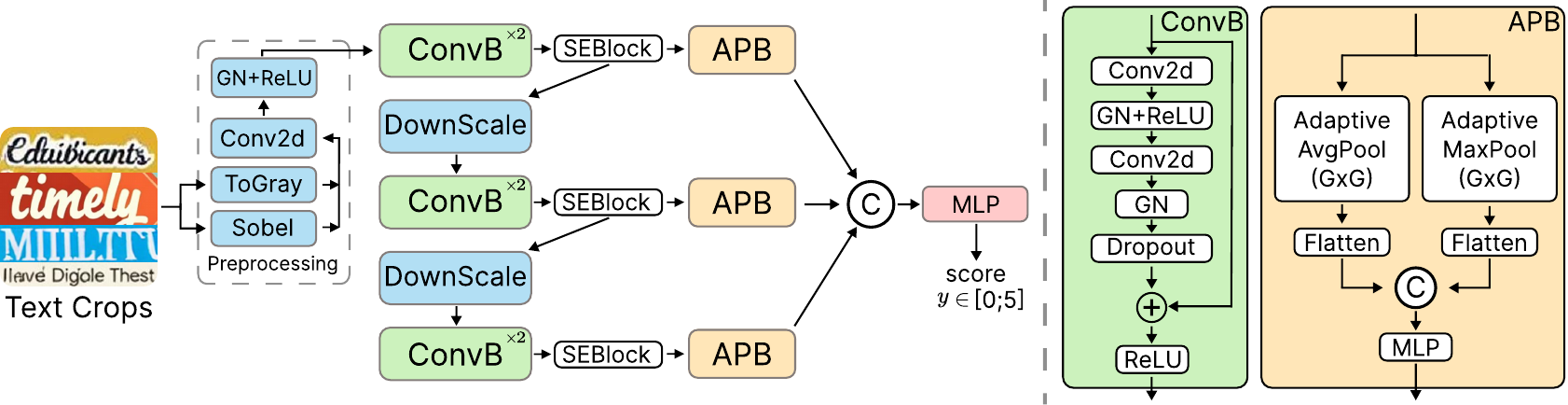}
    \caption{ANTIQA architecture. Each text crop is converted to grayscale, concatenated with a Sobel edge map, and then processed by a lightweight multi-scale CNN with residual stages and downsampling. Features from multiple resolutions are pooled to fixed grids using adaptive average and max pooling, fused via an MLP head, and regressed to a single MOS prediction.}
    \label{fig:antiqa}
    \Description{TODO}
\end{figure*}

\textbf{Image generation.}
We created a set of 30 text-heavy prompts that reliably produce challenging typography (e.g., dense layouts, small fonts, mixed font styles, long paragraphs, numbers, and structured text such as lists or pseudo-documents). We rendered each prompt with 10 recent text-to-image generators via \texttt{replicate.com}, including GPT Image 1.5, Nano Banana Pro, Flux 2 [max], SeeDream 4.5, etc.
For each (model, prompt) pair, we generated 5 images using different random seeds, resulting in total of 1{,}500 images.

\textbf{Text-only rendering.}
For each image, we derive a text-only version that preserves the rendered text while removing surrounding content. Concretely, we detect text regions using PP-OCRv5~\cite{paddle} and construct a binary mask; pixels outside the mask are set to a uniform white background, while pixels inside the text regions are preserved exactly. This isolates the text's perceptual quality from non-textual visual factors.

\textbf{Subjective study protocol.}
We collect human judgments under two complementary rating tasks, each using an integer 0--5 scale (higher is better), and compute the mean opinion score (MOS) as the mean rating across raters:

\textbf{(i) Overall quality (OQ-MOS; full-frame image).} Raters score the overall perceptual quality of the complete image, considering any visible degradations (e.g., blur, noise, and text artifacts).

\textbf{(ii) Text quality (TQ-MOS; text-only).} Raters score only the \emph{perceptual quality of the text}, using the corresponding text-only image. They are instructed to ignore semantics (meaning, correctness, or sense of the written content) and judge only visual artifacts such as malformed glyphs, broken strokes, character substitutions, spacing/kerning issues, and inconsistent baselines.
The two tasks are run independently, yielding paired MOS annotations for overall image quality and text-only quality. Full list of used T2I models and their parameters, examples of images from the datasets, curated list of prompts for TIQA-Images, and subjective instructions can be found in Appendix~\ref{app:human}. We also evaluate participants’ ability to separate visual quality from semantics (Appendix~\ref{app:human}).

\section{AI-generated No-reference Text-in-Image Quality Assessment (ANTIQA) model}

We design the architecture and training procedure with the following criteria in mind: (i) the model must capture fine-grained glyph details and global word-level structure; (ii) the model should be robust across fonts/styles/generators; (iii) the model should be fast enough for large-scale use.

\subsection{Architecture}
As shown in Figure~\ref{fig:antiqa}, ANTIQA predicts a single MOS score from a text crop represented by a 2-channel input (grayscale concatenated with a Sobel edge map). A lightweight stem projects the input to 64 channels, after which the network proceeds through three resolution stages of repeated \textbf{ConvB} blocks separated by two \textbf{DownScale} modules that halve spatial size and double the channel count ($64 \to 128 \to 256$). At the end of each stage, a Squeeze-and-Excitation gate~\cite{Hu_2018_CVPR} recalibrates channels and an Adaptive Pooling Block (\textbf{APB}) produces a fixed-size scale embedding. The three per-scale embeddings are concatenated (operator $C$) and passed to a final MLP head that regresses the MOS score $y \in [0,5]$.

\textbf{ConvB} (Figure~\ref{fig:antiqa}, right) is a residual block of two $3{\times}3$ convolutions with GroupNorm, an inner ReLU, Dropout2d on the residual branch, and a final ReLU after the skip connection. We use GroupNorm rather than BatchNorm because batches of text crops are statistically heterogeneous (mixed fonts, scripts, and degradations), conditions under which GN is more stable. \textbf{APB} extracts features at each scale via parallel adaptive average and max pooling to a $G{\times}G$ grid, projecting their concatenation through a per-scale linear layer to a 64-dimensional embedding: average pooling captures the dominant channel response while max pooling preserves localized high-activation evidence (e.g., a single severely degraded glyph), making the two complementary for quality regression. ANTIQA contains 3.8M parameters and requires 31.5 GFLOPs per $256{\times}256$ crop, enabling efficient evaluation. Full layer-by-layer specifications are provided in Appendix~\ref{app:impl:arch}.

\begin{table*}[tb!]
    \centering
    \small
    \setlength{\tabcolsep}{7pt}
    \renewcommand{\arraystretch}{1.15}
    \caption{Performance on TIQA-Crops (crop-level) and TIQA-Images (image-level) measured by PLCC/SROCC with human MOS. 
    ``$^{\star}$'' denotes finetuned IQA models. Speed is computed on $256\times256$ images on NVIDIA A100 GPU. For VLMs, ``A$x$B'' denotes $x$ active parameters out of the MoE total.}
    \begin{tabular}{llcccccccc} 
        \toprule
        \multirow{2}{*}{Type} & \multirow{2}{*}{Model} & \multicolumn{2}{c}{TIQA-Crops} & \multicolumn{2}{c}{TIQA-Images (OQ-MOS)} & \multicolumn{2}{c}{TIQA-Images (TQ-MOS)} & \multirow{2}{*}{Params} & \multirow{2}{*}{Speed (FPS)}\\
        \cmidrule(lr){3-4}\cmidrule(lr){5-6}\cmidrule(lr){7-8}
        & & PLCC$\uparrow$ & SROCC$\uparrow$ & PLCC$\uparrow$ & SROCC$\uparrow$ &  PLCC$\uparrow$ & SROCC$\uparrow$ & & \\
        \midrule
        \multirow{2}{*}{\makecell[l]{Generic\\supervised models}} 
            & ResNet50 & 0.917 & 0.920 & 0.735 & 0.732 & 0.728 & 0.731 & 25.6\,M & \underline{220.4} \\
            & ViT & \underline{0.926} & \underline{0.927} & 0.740 & 0.738 & 0.734 & 0.735 & 86.6\,M & \textbf{244.7} \\
        \addlinespace
        \multirow{4}{*}{\textsc{IQA}} 
            & TOPIQ & 0.401 & 0.414 & 0.615 & 0.568 & 0.493 & 0.470 & 45.2\,M & 66.7 \\
            & TOPIQ$^{\star}$ & 0.870 & 0.879 & \underline{0.752} & \underline{0.754} & 0.748 & 0.749 & 45.2\,M & 66.7 \\
            & HyperIQA & 0.622 & 0.668 & 0.607 & 0.592 & 0.501 & 0.497 & 27.4\,M & 97.4 \\
            & HyperIQA$^{\star}$ & 0.861 & 0.875 & 0.750 & 0.746 & 0.743 & 0.739 & 27.4\,M & 97.4 \\
        \addlinespace
        \multirow{4}{*}{\textsc{OCR}} 
            & PaddleOCR & 0.778 & 0.788 & 0.671 & 0.664 & \underline{0.761} & \underline{0.787} & \underline{5.0}\,M & 113.3 \\
            & EasyOCR & 0.699 & 0.737 & 0.640 &  0.636 & 0.681 & 0.695 & {$\sim$}25\,M & 109.1 \\
            & RapidOCR & 0.783 & 0.816 & 0.582 & 0.589 &  0.668 & 0.653 & {$\sim$}10\,M & 126.7 \\
            & SAR & 0.690 & 0.709 & 0.569 & 0.591 & 0.634 & 0.640 & {$\sim$}27\,M & 19.1\\
        \addlinespace
        \multirow{2}{*}{\textsc{VLM}} 
            & Qwen3-VL & 0.891 & 0.921 & 0.471 & 0.443 & 0.447 & 0.424 & 235\,B/A22\,B & 0.6 \\
            & GLM-4.6V & 0.674 & 0.671 & 0.257 & 0.343 & 0.193 & 0.288 & 106\,B/A12\,B & 0.4 \\
        \addlinespace
        \makecell[l]{\textsc{TIQA}} 
            & ANTIQA (ours) & \textbf{0.942} & \textbf{0.935} & \textbf{0.810} & \textbf{0.797} & \textbf{0.842} & \textbf{0.837} & \textbf{3.8\,M} &
            119.0\\
        \bottomrule
    \end{tabular}
    \label{tab:main_results}
\end{table*}

\subsection{Training}
We first pretrain ANTIQA on 110k text crops without MOS scores using OCR confidence scores from the PP-OCRv5 model mapped to the MOS range. The OCR confidence\(\rightarrow\)MOS mapping is computed via neural optimal transport~\cite{korotin2022neural}, aligning the proxy-score distribution to the MOS distribution while preserving monotonicity in practice. To compute the mapping function we used only the training split of the TIQA-Crops. These 110k crops are disjoint (by image ID) from the 10k MOS-labeled crops. We then finetune on 10,000 MOS-labeled crops from TIQA-Crops dataset with a mixed objective combining MSE and pairwise ordering: $\mathcal{L}=\mathcal{L}_{\mathrm{MSE}}+\lambda\,\mathcal{L}_{\mathrm{rank}}$:
\begin{equation}
\begin{aligned}
\mathcal{L}_{\mathrm{MSE}} &= \frac{1}{B}\sum_{i=1}^{B}(y_i-\hat{y}_i)^2, \\
\mathcal{L}_{\mathrm{rank}} &= \frac{1}{|B|}\sum_{i<j}
\left[\mathrm{softplus}\!\left(-\mathrm{sign}(y_i-y_j)(\hat{y}_i-\hat{y}_j)\right)\right],
\end{aligned}
\end{equation}
where $y$ denotes MOS value, $\hat{y}$ is predicted score, and $|B|$ is the size of a mini-batch. 
This encourages both calibrated scores and correct relative preferences, matching correlation-based evaluation.
Architecture, training details and an ablation study for ANTIQA’s design choices are provided in Appendix~\ref{app:results}.

\section{Experiments}

\subsection{Experimental Setup}

For evaluation, we employ two widely used correlation coefficients for MOS-annotated quality assessment: Pearson’s Linear Correlation Coefficient (PLCC) and Spearman’s Rank Order Correlation Coefficient (SROCC).

To prevent leakage from near-duplicate crops, we split TIQA-Crops by source image ID (all crops from the same image are assigned to the same split). We report results only on the held-out test split. This way, the 10,000 MOS-annotated crops from TIQA-Crops were split into training (9,000), validation (500), and test (500) sets. TIQA-Images was used in its entirety without further splitting.

We compare against four baseline families: (i) OCR confidence scores: PaddleOCR~3.0 (PP-OCRv5)~\cite{paddle}, EasyOCR~\cite{easyocr}, \\ RapidOCR~\cite{rapidocr}, SAR~\cite{sarocr}, all used out-of-the-box with their default detector–recognizer pipelines; (ii) VLM-based judges: \\ \texttt{Qwen3-VL-235B-A22B-Instruct}~\cite{qwen3} and \texttt{GLM-4.6V}~\cite{glm46}, both \\ Mixture-of-Experts vision–language models (235B/22B-active and 106B/12B-active parameters, respectively) queried via their public APIs; (iii) general no-reference IQA metrics TOPIQ~\cite{topiq} and HyperIQA~\cite{hyperiqa} (loaded from the \texttt{pyiqa} toolbox); and (iv) widely-used backbones ResNet50~\cite{resnet} and ViT~\cite{vit} fine-tuned from ImageNet-pretrained weights. For VLM judges, we prompt the model to score \emph{text rendering fidelity only} on a 0--5 scale (floats allowed), explicitly instructing it to ignore textual meaning and spelling; the score is the first parsed number in the response, and we use a fixed temperature of $0$ to make outputs reproducible. Prompts are provided in Appendix~\ref{app:results}.

To ensure a fair comparison and isolate the contribution of the architectural design, we train all baselines using the same two-stage procedure as ANTIQA: synthetic pretraining on the 110k OCR-pseudo-labeled crops followed by fine-tuning on the 10k MOS-labeled crops, as we found this protocol consistently outperformed direct training on MOS labels. The ResNet50 and ViT backbones are initialized from ImageNet-pretrained weights with a regression head, while TOPIQ and HyperIQA are fine-tuned from their official pretrained checkpoints. Full hyperparameters and training configurations for each baseline are reported in Appendix~\ref{app:impl}.

\textbf{Image-level aggregation for TIQA-Images dataset.}
\label{sec:aggregation}
We aggregate crop scores into an image-level score via area-weighted pooling:
\begin{equation}
Q_{\text{ANTIQA}}(x) = \frac{\sum_{i=1}^{N} w_i\, q(c_i)}{\sum_{i=1}^{N} w_i},
\qquad
w_i = \text{area}(c_i),
\label{eq:tiqa_pool}
\end{equation}
where $c_i$ denotes a text crop from image $x$, $N$ is the number of text crops in $x$ and $q(c_i)$ is the model’s predicted quality for crop $c_i$.
We also report an ablation of alternative pooling strategies in Appendix~\ref{app:impl}.

\begin{table*}[tb!]
    \centering
    \small
    \caption{
    Best-of-$K$ ranking/selection on TIQA-Images.
    Within-group PLCC/SROCC to MOS are averaged over groups (mean$\pm$std).
    Selection reports the mean MOS of the top-scored image per group and $\Delta$ over Random.
    Gap closed is the improvement over Random relative to Oracle. ``$^{\star}$'' denotes finetuned IQA models. ``*'' denotes text-only masked image. ``**'' denotes separate crops as input and averaging their scores. 
    Random is averaged over 1{,}000 runs; Oracle is max value over each group.
    }
    \setlength{\tabcolsep}{4pt}
    \begin{tabular}{llcccccccc}
        \toprule
        \multirow{2}{*}{Type} & \multirow{2}{*}{Model}
        & \multicolumn{4}{c}{Within-group correlation to MOS (mean (std)) $\uparrow$}
        & \multicolumn{4}{c}{Best-of-5 selection outcome (mean MOS / gain)} \\
        \cmidrule(lr){3-6}\cmidrule(lr){7-10}
        &
        & \multicolumn{2}{c}{TQ-MOS} & \multicolumn{2}{c}{OQ-MOS}
        & \multicolumn{2}{c}{Selected MOS} & \multicolumn{2}{c}{$\Delta$MOS vs Random (gap closed)} \\
        \cmidrule(lr){3-4}\cmidrule(lr){5-6}\cmidrule(lr){7-8}\cmidrule(lr){9-10}
        &
        & PLCC & SROCC & PLCC & SROCC
        & TQ & OQ & TQ & OQ \\
        \midrule

        \multirow{2}{*}{Reference} & Random
          & --- & --- & --- & --- 
          & 2.57 & 3.01 
          & +0.00 {\scriptsize (0\%)} & +0.00 {\scriptsize (0\%)} \\
        & Oracle
          & --- & --- & --- & --- 
          & 3.07 & 3.47 
          & +0.50 {\scriptsize (100\%)} & +0.46 {\scriptsize (100\%)} \\
        \midrule

        \multirow{2}{*}{\makecell[l]{Generic\\supervised models}} 
        & ResNet50
          & 0.351 {\scriptsize (0.112)} & 0.364 {\scriptsize (0.140)}
          & 0.260 {\scriptsize (0.127)} & 0.265 {\scriptsize (0.130)}
          & 2.69 & 3.06
          & +0.12 {\scriptsize (24\%)} & +0.05 {\scriptsize (10.9\%)} \\
        & ViT
          & 0.342 {\scriptsize (0.128)} & 0.359 {\scriptsize (0.119)}
          & 0.274 {\scriptsize (0.113)} & 0.259 {\scriptsize (0.109)}
          & 2.70 & 3.08
          & +0.13 {\scriptsize (26\%)} & +0.07 {\scriptsize (14.1\%)} \\
        \addlinespace
        
        \multirow{4}{*}{IQA} 
        & TOPIQ
          & 0.115 {\scriptsize (0.147)} & 0.096 {\scriptsize (0.151)}
          & \underline{0.296 {\scriptsize (0.125)}} & \underline{0.280 {\scriptsize (0.130)}}
          & 2.71 & 3.14
          & +0.14 {\scriptsize (28\%)} & +0.13 {\scriptsize (28.3\%)} \\
        & TOPIQ$^{\star}$
          & 0.340 {\scriptsize (0.103)} & 0.361 {\scriptsize (0.131)}
          & 0.258 {\scriptsize (0.126)} & 0.263 {\scriptsize (0.117)}
          & 2.75 & 3.09
          & +0.18 {\scriptsize (36\%)} & +0.08 {\scriptsize (17.4\%)} \\
        & HyperIQA
          & 0.134 {\scriptsize (0.109)} & 0.140 {\scriptsize (0.104)}
          & 0.276 {\scriptsize (0.121)} & \underline{0.280 {\scriptsize (0.109)}}
          & 2.72 & 3.09
          & +0.15 {\scriptsize (30\%)} & +0.08 {\scriptsize (17.4\%)} \\
        & HyperIQA$^{\star}$
          & 0.329 {\scriptsize (0.112)} & 0.341 {\scriptsize (0.114)}
          & 0.265 {\scriptsize (0.123)} & 0.276 {\scriptsize (0.119)}
          & 2.74 & 3.07
          & +0.17 {\scriptsize (34\%)} & +0.06 {\scriptsize (13.0\%)} \\
        \addlinespace

        \multirow{2}{*}{OCR} & PaddleOCR
          & \underline{0.415 {\scriptsize (0.132)}} & \underline{0.364 {\scriptsize (0.134)}}
          & 0.171 {\scriptsize (0.161)} & 0.165 {\scriptsize (0.162)}
          & \underline{2.91} & \underline{3.15}
          & \underline{+0.34 {\scriptsize (68\%)}} & \underline{+0.14 {\scriptsize (30.4\%)}} \\
        & SAR
          & 0.325 {\scriptsize (0.145)} & 0.319 {\scriptsize (0.131)}
          & 0.120 {\scriptsize (0.167)} & 0.116 {\scriptsize (0.159)}
          & 2.81 & 3.08
          & +0.24 {\scriptsize (48\%)} & +0.07 {\scriptsize (15.2\%)} \\
        \addlinespace

        \multirow{6}{*}{VLM} & Qwen3
          & 0.060 {\scriptsize (0.176)} & 0.049 {\scriptsize (0.171)}
          & 0.099 {\scriptsize (0.163)} & 0.122 {\scriptsize (0.156)}
          & 2.68 & 3.08
          & +0.11 {\scriptsize (22\%)} & +0.07 {\scriptsize (14.1\%)} \\
        & Qwen3*
          & 0.181 {\scriptsize (0.140)} & 0.162 {\scriptsize (0.129)}
          & 0.131 {\scriptsize (0.158)} & 0.125 {\scriptsize (0.149)}
          & 2.75 & 3.12
          & +0.18 {\scriptsize (36\%)} & +0.11 {\scriptsize (22.8\%)} \\
        & Qwen3**
          & 0.265 {\scriptsize (0.129)} & 0.249 {\scriptsize (0.121)}
          & 0.219 {\scriptsize (0.157)} & 0.211 {\scriptsize (0.146)}
          & 2.81 & 3.13
          & +0.24 {\scriptsize (48\%)} & +0.12 {\scriptsize (26.1\%)} \\
        & GLM 4.6
          & 0.027 {\scriptsize (0.167)} & 0.026 {\scriptsize (0.157)}
          & 0.067 {\scriptsize (0.153)} & 0.052 {\scriptsize (0.149)}
          & 2.66 & 3.06
          & +0.09 {\scriptsize (18\%)} & +0.05 {\scriptsize (10.9\%)} \\
        & GLM 4.6*
          & 0.148 {\scriptsize (0.151)} & 0.136 {\scriptsize (0.149)}
          & 0.112 {\scriptsize (0.153)} & 0.102 {\scriptsize (0.151)}
          & 2.74 & 3.10
          & +0.17 {\scriptsize (35\%)} & +0.09 {\scriptsize (18.5\%)} \\
        & GLM 4.6**
          & 0.185 {\scriptsize (0.149)} & 0.174 {\scriptsize (0.147)}
          & 0.136 {\scriptsize (0.155)} & 0.142 {\scriptsize (0.154)}
          & 2.76 & 3.12
          & +0.19 {\scriptsize (38\%)} & +0.11 {\scriptsize (23.9\%)} \\
        \addlinespace
        
        TIQA & ANTIQA
          & \textbf{0.419 {\scriptsize (0.112)}} & \textbf{0.382 {\scriptsize (0.111)}}
          & \textbf{0.388 {\scriptsize (0.130)}} & \textbf{0.340 {\scriptsize (0.135)}}
          & \textbf{2.93} & \textbf{3.31}
          & \textbf{+0.36 {\scriptsize (72\%)}} & \textbf{+0.30 {\scriptsize (65.2\%)}} \\
        \bottomrule
    \end{tabular}
    \label{tab:ranking}
\end{table*}

\subsection{Results on TIQA-Crops}
Table~\ref{tab:main_results} summarizes correlation coefficients on the TIQA-Crops test set for ANTIQA and other baseline models. 
On TIQA-Crops, ANTIQA outperforms OCR confidence and VLM judges on crop-level MOS, indicating that recognizing text is not sufficient: the model must also capture visual degradations specific to rendered glyphs (e.g., stroke breaks, bleeding, aliasing). The relatively strong Qwen3 crop performance suggests VLMs can judge local text when text occupies most pixels, but this advantage does not directly transfer to full-image scoring on TIQA-Images. VLM models are also computationally heavy, with an average FPS of 0.6 for Qwen3.
In contrast, off-the-shelf general NR-IQA baselines perform substantially worse than text-specific signals, while finetuning markedly improves them; nevertheless, even the strongest finetuned generic IQA models (marked with $^{\star}$) remain below ANTIQA, indicating that general-purpose IQA still does not fully capture rendered-text quality. The large gap between off-the-shelf and finetuned IQA models shows that generic image-quality features are not useless for TIQA, but without text-focused adaptation they are poorly aligned with the artifact types that humans penalize in rendered text.
Generic supervised models ResNet50 and ViT trained from scratch perform relatively strong, but lack text-specific inductive biases that make ANTIQA superior.


\subsection{Results on TIQA-Images}
We evaluate whether a dedicated text-in-image quality assessor (ANTIQA) better matches human judgments than generic image-quality evaluation methods on \emph{unseen} modern text-heavy T2I outputs. None of the 10 generative models represented in TIQA-Images was used during ANTIQA training. As described in Section~\ref{sec:TIQA_images}, TIQA-Images dataset contains OQ-MOS and TQ-MOS for each image, corresponding to overall and text quality, respectively. Crop-based methods (ANTIQA, OCR, IQA, generic supervised models) operate on detected text regions and are pooled to image-level, as described in Eq.~\ref{eq:tiqa_pool}; full-frame baselines (VLM judges) score the entire image for Table~\ref{tab:main_results}. We also evaluate VLM judges under three different computation modes in Table~\ref{tab:ranking}. 

\textbf{ANTIQA generalizes well to unseen SOTA generators.}
We evaluate overall alignment with humans across all 1,500 images, reporting SROCC and PLCC between method scores and MOS values.
Table~\ref{tab:main_results} reports global correlations on TIQA-Images for both overall and text-only MOS (OQ-MOS and TQ-MOS), where ANTIQA achieves the strongest correlation with both. Among non-specialized baselines, the finetuned IQA models are strongest overall, with the generic supervised backbones ViT and ResNet50 also competitive; yet ANTIQA remains clearly ahead on both OQ-MOS and TQ-MOS. Notably, TIQA-Images contains images from T2I models unseen during ANTIQA training, indicating good generalization to unseen SOTA generators.

\textbf{VLM judging is localization-sensitive.}
Table~\ref{tab:ranking} also reveals that VLM judges are substantially stronger when their input is restricted to the text region. On full images, rendered text often occupies few pixels, so the judgment can be dominated by non-text content, diluting sensitivity to glyph-level artifacts. The text-only masked view (marked by *) reduces this context leakage by suppressing non-text regions, improving alignment with TQ-MOS. Evaluating the VLM per detected text crop and averaging across crops (marked by **) further helps by (i) zooming in to preserve stroke-level details and (ii) reducing variance by aggregating multiple local judgments into a stable image-level score.

\textbf{Text and overall quality scores are strongly coupled.}
Text and overall quality scores are strongly coupled. We observe a strong association between overall quality (OQ-MOS) and text quality (TQ-MOS) on TIQA-Images (SROCC $\approx 0.78$ over the full dataset). To verify this is not purely an across-generator effect, we decompose the correlation by the dataset hierarchy (Appendix, Table~\ref{tab:iq_tq_decomposition}). Across generator-level averages (10 groups), it is near-perfect (PLCC=0.96, SROCC=0.98); more importantly, holding generator and prompt fixed and varying only the random seed (5 images per pair), the within-group relationship stays strongly positive (mean/median SROCC 0.51/0.59). Thus the OQ-TQ coupling is not merely driven by differences between generators but persists within fixed generator–prompt settings, implying that in this text-heavy regime overall preference is largely constrained by text-rendering failure — consistent with a text-specialized signal outperforming generic NR-IQA on OQ-MOS despite being trained for text quality.

\textbf{ANTIQA excels at best-of-$K$ selection.}
To test whether a method can identify the best sample among generations of the \emph{same} prompt and generator, we evaluate ranking on each group of $K{=}5$ images from TIQA-Images, where a group is one (prompt, generator) pair. Per group we correlate predicted scores with MOS across the $K$ samples, then average over all $S{=}300$ groups.
We report both TQ-MOS and OQ-MOS; this is especially hard here, as small within-group differences must be detected across only $K{=}5$ samples.
Table~\ref{tab:ranking} shows ANTIQA achieves the strongest within-group agreement with human rankings for both TQ-MOS (PLCC/SROCC: 0.419/0.382) and OQ-MOS (0.388/0.340).
Generic NR-IQA (e.g., TOPIQ) correlates better with OQ than TQ, consistent with IQA emphasizing global naturalness over text legibility.
OCR confidence (PaddleOCR) is competitive for text (0.415/0.364) but transfers poorly to overall quality (0.171/0.165): recognizing text does not capture visual preference.
VLM-based scorers underperform on both, suggesting limited calibration for fine-grained, within-prompt comparisons.
ResNet50 and ViT give only moderate within-group accuracy; finetuning generic IQA models substantially improves text-quality ranking over off-the-shelf versions, yet ANTIQA still leads. Notably, comparing Tables~\ref{tab:main_results} and~\ref{tab:ranking} shows that strong global correlation does not automatically yield strong best-of-5 ranking: the generic backbones look competitive on average but are much weaker at distinguishing subtle seed-level differences under a fixed prompt and generator.

\textbf{Best-of-$K$ selection improves human MOS.}
\label{sec:reranking}
Correlation captures ranking consistency, but in practice, we aim to select the best sample based on quality. For each (prompt, generator) group, we choose the image with the highest predicted quality score produced by a given method and report its average MOS:
\begin{equation}
\mathrm{MOS}^{\uparrow}
\;=\;
\frac{1}{S}\sum_{i=1}^{S}
\mathrm{MOS}\!\left(s_i^{k_i^{*}}\right), k_i^{*}=\arg\max_{k\in\{1,\dots,K\}} q_i^{(k)}
\label{eq:best_of_k}
\end{equation}
where $\{s_i^{1},\ldots,s_i^{K}\}$ is the set of $K$ samples in group $i$, and
$q_i^k$ is the corresponding predicted score output by the evaluated method.
The selected index $k_i^{*}$ corresponds to the sample that the method predicts to be best within group $s_i$.

As shown in Table~\ref{tab:ranking}, ANTIQA yields the largest MOS gains over Random selection: $+0.36$ on TQ-MOS (2.57 $\rightarrow$ 2.93, 14\% improvement) and $+0.30$ on OQ-MOS (3.01 $\rightarrow$ 3.31, 9.9\% improvement), closing \textbf{72\%} and \textbf{65.2\%} of the gap to an Oracle selector, respectively.
Notably, PaddleOCR improves TQ-MOS substantially (+0.34) but provides only a modest OQ-MOS gain (+0.14), reinforcing that OCR confidence captures legibility but misses broader factors that drive overall human preference. In summary, ANTIQA consistently selects higher-MOS images, making it a strong drop-in signal for generation-time filtering and reranking when prompts and generators are held fixed. We also evaluate the ability of ANTIQA to predict failures of OCR and VLM vision tasks in Appendix~\ref{app:results}.

\subsection{Text quality comparison of the latest T2I models}

\begin{figure}[t!]
    \vspace{-6pt}
    \centering
    \includegraphics[width=\linewidth]{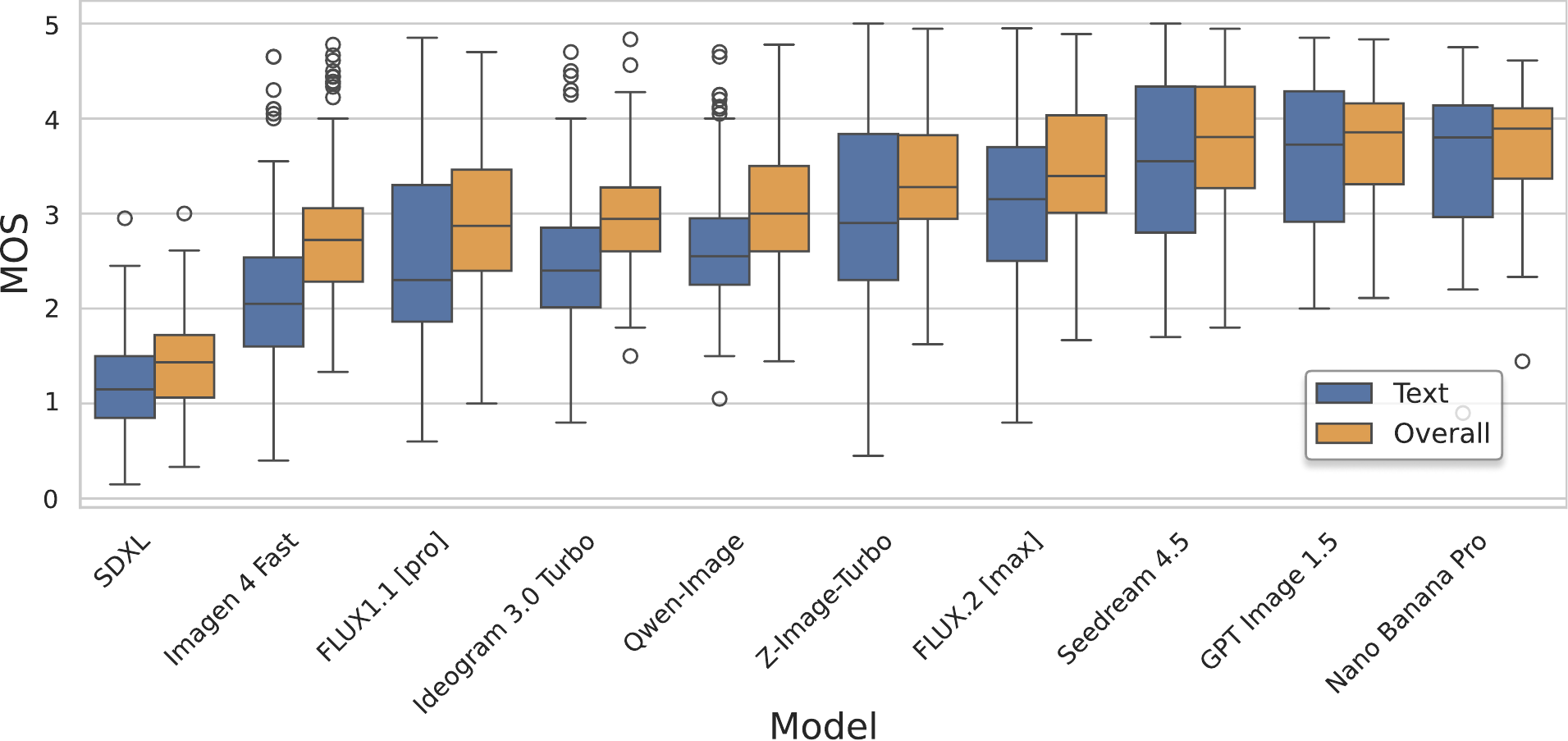}
    \caption{Box-plot distributions of OQ-MOS and TQ-MOS for separate generators. The models are sorted by mean TQ-MOS.}
    \label{fig:t2i_comp}
    \Description{TODO}
\end{figure}

TIQA-Images was also used to analyze text-rendering quality in recent T2I models. As described in Section~\ref{sec:TIQA_images}, we collected MOS for the text-only version of AI-generated images; Figure~\ref{fig:t2i_comp} shows box-plots of both MOS distributions per generator, making explicit that overall and rendered-text quality, though correlated, are not interchangeable and should be evaluated independently. Text rendering improves clearly from older baselines (e.g., SDXL) to the newest models, yet OQ consistently exceeds TQ for every generator: images can look convincing while the embedded text remains degraded. Robustness remains the main limitation—TQ distributions show low-score outliers, reflecting rare but severe failures and strong sensitivity to prompts and rendering; thus, when text quality is critical, tail behavior is more informative than the median.

\section{Conclusion}
We argue that perceptual quality of rendered text in generated images is a distinct evaluation problem, not a by-product of OCR correctness or generic image-quality judgment. We formalize it as Text-in-Image Quality Assessment (TIQA) and make it measurable through two complementary datasets spanning crop-level and full-image evaluation on recent text-heavy generations. On this benchmark, ANTIQA achieves the strongest alignment with human judgments—outperforming OCR-based scores, generic backbones (ResNet50, ViT), NR-IQA models including finetuned variants, and VLM-based judges, including on unseen generators. It is also practical: ANTIQA improves best-of-5 selection for both text-only and overall image MOS, an effective control signal for filtering and reranking. More broadly, rendered text is not a minor defect but a major driver of human preference, and persistent tail failures leave typography an unresolved bottleneck.


\begin{acks}
The work of Aleksandr Guschin and Anastasia Antsiferova were supported by a grant, provided by the Ministry of Economic Development of the Russian Federation (agreement  dated June 20, 2025 No. 139-15-2025-011, identifier 000000C313925P4G0002). 
The research was carried out using the MSU-270 supercomputer of Lomonosov Moscow State University and ISP RAS computing cluster. 
\end{acks}

\clearpage



\bibliographystyle{ACM-Reference-Format}
\bibliography{sample-base}

@String{Computing = "Computing" }

@String{Computer = "{IEEE} Computer" }

@INPROCEEDINGS{rank,
  author={Liu, Xialei and Van De Weijer, Joost and Bagdanov, Andrew D.},
  booktitle={2017 IEEE International Conference on Computer Vision (ICCV)}, 
  title={RankIQA: Learning from Rankings for No-Reference Image Quality Assessment}, 
  year={2017},
  volume={},
  number={},
  pages={1040-1049},
  doi={10.1109/ICCV.2017.118}}

@inproceedings{deqa_score,
    title={Teaching Large Language Models to Regress Accurate Image Quality Scores using Score Distribution},
    author={You, Zhiyuan and Cai, Xin and Gu, Jinjin and Xue, Tianfan and Dong, Chao},
    booktitle={IEEE/CVF Conference on Computer Vision and Pattern Recognition},
    pages={14483--14494},
    year={2025}
}

@article{korotin2022neural,
  title={Neural optimal transport},
  author={Korotin, Alexander and Selikhanovych, Daniil and Burnaev, Evgeny},
  journal={arXiv preprint arXiv:2201.12220},
  year={2022}
}

@inproceedings{Hu_2018_CVPR,
  author    = {Hu, Jie and Shen, Li and Sun, Gang},
  title     = {Squeeze-and-Excitation Networks},
  booktitle = {Proceedings of the IEEE/CVF Conference on Computer Vision and Pattern Recognition (CVPR)},
  year      = {2018},
  month     = {June},
  pages     = {7132--7141},
  doi       = {10.1109/CVPR.2018.00745}
}

@article{eyring2024reno,
  title={Reno: Enhancing one-step text-to-image models through reward-based noise optimization},
  author={Eyring, Luca and Karthik, Shyamgopal and Roth, Karsten and Dosovitskiy, Alexey and Akata, Zeynep},
  journal={Advances in Neural Information Processing Systems},
  volume={37},
  pages={125487--125519},
  year={2024}
}

@article{lee2023aligning,
  title={Aligning text-to-image models using human feedback},
  author={Lee, Kimin and Liu, Hao and Ryu, Moonkyung and Watkins, Olivia and Du, Yuqing and Boutilier, Craig and Abbeel, Pieter and Ghavamzadeh, Mohammad and Gu, Shixiang Shane},
  journal={arXiv preprint arXiv:2302.12192},
  year={2023}
}

@article{image_reward,
  title={Imagereward: Learning and evaluating human preferences for text-to-image generation},
  author={Xu, Jiazheng and Liu, Xiao and Wu, Yuchen and Tong, Yuxuan and Li, Qinkai and Ding, Ming and Tang, Jie and Dong, Yuxiao},
  journal={Advances in Neural Information Processing Systems},
  volume={36},
  pages={15903--15935},
  year={2023}
}

@article{pickscore,
  title={Pick-a-pic: An open dataset of user preferences for text-to-image generation},
  author={Kirstain, Yuval and Polyak, Adam and Singer, Uriel and Matiana, Shahbuland and Penna, Joe and Levy, Omer},
  journal={Advances in neural information processing systems},
  volume={36},
  pages={36652--36663},
  year={2023}
}

@article{bosheah2025challenges,
  title={Challenges in Generating Accurate Text in Images: A Benchmark for Text-to-Image Models on Specialized Content},
  author={Bosheah, Zenab and Bilicki, Vilmos},
  journal={Applied Sciences},
  volume={15},
  number={5},
  pages={2274},
  year={2025},
  publisher={MDPI}
}

@article{sampaio2024typescore,
  title={Typescore: A text fidelity metric for text-to-image generative models},
  author={Sampaio, Georgia Gabriela and Zhang, Ruixiang and Zhai, Shuangfei and Gu, Jiatao and Susskind, Josh and Jaitly, Navdeep and Zhang, Yizhe},
  journal={arXiv preprint arXiv:2411.02437},
  year={2024}
}

@inproceedings{depictqa_v1,
    title={Depicting Beyond Scores: Advancing Image Quality Assessment through Multi-modal Language Models},
    author={You, Zhiyuan and Li, Zheyuan and Gu, Jinjin and Yin, Zhenfei and Xue, Tianfan and Dong, Chao},
    booktitle={European Conference on Computer Vision},
    pages={259--276},
    year={2024}
}

@inproceedings{
zhu2024adaptive,
title={Adaptive Image Quality Assessment via Teaching Large Multimodal Model to Compare},
author={Hanwei Zhu and Haoning Wu and Yixuan Li and Zicheng Zhang and Baoliang Chen and Lingyu Zhu and Yuming Fang and Guangtao Zhai and Weisi Lin and Shiqi Wang},
booktitle={The Thirty-eighth Annual Conference on Neural Information Processing Systems},
year={2024},
url={https://openreview.net/forum?id=mHtOyh5taj}
}

@article{gu2024survey,
	title   = {A Survey on LLM-as-a-Judge},
	author  = {Jiawei Gu and Xuhui Jiang and Zhichao Shi and Hexiang Tan and Xuehao Zhai and Chengjin Xu and Wei Li and Yinghan Shen and Shengjie Ma and Honghao Liu and Yuanzhuo Wang and Jian Guo},
	year    = {2024},
	journal = {arXiv preprint arXiv: 2411.15594}
}

@inproceedings{postedit_placeholder,
  title={Type-r: Automatically retouching typos for text-to-image generation},
  author={Shimoda, Wataru and Inoue, Naoto and Haraguchi, Daichi and Mitani, Hayato and Uchida, Seiichi and Yamaguchi, Kota},
  booktitle={Proceedings of the Computer Vision and Pattern Recognition Conference},
  pages={2745--2754},
  year={2025}
}

@article{text_guidance_placeholder,
  title={Textdiffuser: Diffusion models as text painters},
  author={Chen, Jingye and Huang, Yupan and Lv, Tengchao and Cui, Lei and Chen, Qifeng and Wei, Furu},
  journal={Advances in Neural Information Processing Systems},
  volume={36},
  pages={9353--9387},
  year={2023}
}

@article{textgen_placeholder,
  title={Anytext: Multilingual visual text generation and editing},
  author={Tuo, Yuxiang and Xiang, Wangmeng and He, Jun-Yan and Geng, Yifeng and Xie, Xuansong},
  journal={arXiv preprint arXiv:2311.03054},
  year={2023}
}

@article{legibility_placeholder,
  title={Legibility of texts: The influence of blur},
  author={Colombo, EM and Kirschbaum, CF and Raitelli, M},
  journal={Lighting Research \& Technology},
  volume={19},
  number={3},
  pages={61--71},
  year={1987},
  publisher={SAGE Publications Sage UK: London, England}
}

@article{schuhmann2022laion5b,
  title={Laion-5b: An open large-scale dataset for training next generation image-text models},
  author={Schuhmann, Christoph and Beaumont, Romain and Vencu, Richard and Gordon, Cade and Wightman, Ross and Cherti, Mehdi and Coombes, Theo and Katta, Aarush and Mullis, Clayton and Wortsman, Mitchell and others},
  journal={Advances in neural information processing systems},
  volume={35},
  pages={25278--25294},
  year={2022}
}

@article{sciqa_placeholder,
  title={Screen content quality assessment: Overview, benchmark, and beyond},
  author={Min, Xiongkuo and Gu, Ke and Zhai, Guangtao and Yang, Xiaokang and Zhang, Wenjun and Le Callet, Patrick and Chen, Chang Wen},
  journal={ACM Computing Surveys (CSUR)},
  volume={54},
  number={9},
  pages={1--36},
  year={2021},
  publisher={ACM New York, NY}
}

@inproceedings{diqa_placeholder,
  title={Document image quality assessment: A brief survey},
  author={Ye, Peng and Doermann, David},
  booktitle={2013 12th International Conference on Document Analysis and Recognition},
  pages={723--727},
  year={2013},
  organization={IEEE}
}

@article{zhang2025astrict,
  title={STRICT: Stress Test of Rendering Images Containing Text},
  author={Zhang, Tianyu and Wang, Xinyu and Tai, Zhenghan and Li, Lu and Chi, Jijun and Tian, Jingrui and He, Hailin and Wang, Suyuchen},
  journal={arXiv preprint arXiv:2505.18985},
  year={2025}
}

@inproceedings{textinvision,
  title={Textinvision: Text and prompt complexity driven visual text generation benchmark},
  author={Fallah, Forouzan and Patel, Maitreya and Chatterjee, Agneet and Morariu, Vlad and Baral, Chitta and Yang, Yezhou},
  booktitle={Proceedings of the IEEE/CVF Conference on Computer Vision and Pattern Recognition (CVPR) Workshops},
  pages={525--534},
  year={2025}
}

@misc{paddle,
      title={PaddleOCR 3.0 Technical Report}, 
      author={Cheng Cui and Ting Sun and Manhui Lin and Tingquan Gao and Yubo Zhang and Jiaxuan Liu and Xueqing Wang and Zelun Zhang and Changda Zhou and Hongen Liu and Yue Zhang and Wenyu Lv and Kui Huang and Yichao Zhang and Jing Zhang and Jun Zhang and Yi Liu and Dianhai Yu and Yanjun Ma},
      year={2025},
      eprint={2507.05595},
      archivePrefix={arXiv},
      primaryClass={cs.CV},
      url={https://arxiv.org/abs/2507.05595}, 
}

@inproceedings{sarocr,
author = {Li, Hui and Wang, Peng and Shen, Chunhua and Zhang, Guyu},
title = {Show, attend and read: a simple and strong baseline for irregular text recognition},
year = {2019},
isbn = {978-1-57735-809-1},
publisher = {AAAI Press},
url = {https://doi.org/10.1609/aaai.v33i01.33018610},
doi = {10.1609/aaai.v33i01.33018610},
booktitle = {Proceedings of the Thirty-Third AAAI Conference on Artificial Intelligence and Thirty-First Innovative Applications of Artificial Intelligence Conference and Ninth AAAI Symposium on Educational Advances in Artificial Intelligence},
articleno = {1056},
numpages = {8},
location = {Honolulu, Hawaii, USA},
series = {AAAI'19/IAAI'19/EAAI'19}
}

@misc{qwen3,
      title={Qwen3-VL Technical Report}, 
      author={Shuai Bai and Yuxuan Cai and Ruizhe Chen and Keqin Chen and Xionghui Chen and Zesen Cheng and Lianghao Deng and Wei Ding and Chang Gao and Chunjiang Ge and Wenbin Ge and Zhifang Guo and Qidong Huang and Jie Huang and Fei Huang and Binyuan Hui and Shutong Jiang and Zhaohai Li and Mingsheng Li and Mei Li and Kaixin Li and Zicheng Lin and Junyang Lin and Xuejing Liu and Jiawei Liu and others},
      year={2025},
      eprint={2511.21631},
      archivePrefix={arXiv},
      primaryClass={cs.CV},
      url={https://arxiv.org/abs/2511.21631}, 
}

@misc{glm46,
      title={GLM-4.5V and GLM-4.1V-Thinking: Towards Versatile Multimodal Reasoning with Scalable Reinforcement Learning}, 
      author={V Team and Wenyi Hong and Wenmeng Yu and Xiaotao Gu and Guo Wang and Guobing Gan and Haomiao Tang and Jiale Cheng and Ji Qi and Junhui Ji and Lihang Pan and Shuaiqi Duan and Weihan Wang and Yan Wang and Yean Cheng and Zehai He and Zhe Su and Zhen Yang and Ziyang Pan and Aohan Zeng and Baoxu Wang and Bin Chen and Boyan Shi and Changyu Pang and Chenhui Zhang and Da Yin and Fan Yang and Guoqing Chen and Haochen Li and Jiale Zhu and Jiali Chen and Jiaxing Xu and Jiazheng Xu and Jing Chen and Jinghao Lin and Jinhao Chen and Jinjiang Wang and Junjie Chen and Leqi Lei and Letian Gong and Leyi Pan and Mingdao Liu and others},
      year={2026},
      eprint={2507.01006},
      archivePrefix={arXiv},
      primaryClass={cs.CV},
      url={https://arxiv.org/abs/2507.01006}, 
}

@article{heusel2017gans,
  title={Gans trained by a two time-scale update rule converge to a local nash equilibrium},
  author={Heusel, Martin and Ramsauer, Hubert and Unterthiner, Thomas and Nessler, Bernhard and Hochreiter, Sepp},
  journal={Advances in neural information processing systems},
  volume={30},
  year={2017}
}

@article{topiq,
  title={TOPIQ: A Top-Down Approach From Semantics to Distortions for Image Quality Assessment},
  author={Chen, Chaofeng and Mo, Jiadi and Hou, Jingwen and Wu, Haoning and Liao, Liang and Sun, Wenxiu and Yan, Qiong and Lin, Weisi},
  journal={IEEE Transactions on Image Processing},
  volume={33},
  pages={2404--2418},
  year={2024},
  doi={10.1109/TIP.2024.3378466},
  url={https://doi.org/10.1109/TIP.2024.3378466}
}

@misc{yandex_tasks,
  author       = {{Yandex}},
  title        = {Yandex.Tasks},
  year         = {n.d.},
  howpublished = {\url{https://tasks.yandex.com}},
  note         = {Accessed 20 December 2025}
}

@misc{lmarena,
  author       = {{UC Berkeley}},
  title        = {LMArena},
  year         = {n.d.},
  howpublished = {\url{https://lmarena.ai/leaderboard/text-to-image}},
}

@inproceedings{li2024aigiqa,
  title={Aigiqa-20k: A large database for ai-generated image quality assessment},
  author={Li, Chunyi and Kou, Tengchuan and Gao, Yixuan and Cao, Yuqin and Sun, Wei and Zhang, Zicheng and Zhou, Yingjie and Zhang, Zhichao and Zhang, Weixia and Wu, Haoning and others},
  booktitle={In Proceedings of the IEEE/CVF Conference on Computer Vision and Pattern Recognition (CVPR) Workshops},
  pages={6327--6336},
  year={2024}
}

@article{salimans2016improved,
  title={Improved techniques for training gans},
  author={Salimans, Tim and Goodfellow, Ian and Zaremba, Wojciech and Cheung, Vicki and Radford, Alec and Chen, Xi},
  journal={Advances in neural information processing systems},
  volume={29},
  year={2016}
}

@inproceedings{zhang2018unreasonable,
  title={The unreasonable effectiveness of deep features as a perceptual metric},
  author={Zhang, Richard and Isola, Phillip and Efros, Alexei A and Shechtman, Eli and Wang, Oliver},
  booktitle={Proceedings of the IEEE conference on computer vision and pattern recognition},
  pages={586--595},
  year={2018}
}

@article{mittal2012no,
  title={No-reference image quality assessment in the spatial domain},
  author={Mittal, Anish and Moorthy, Anush Krishna and Bovik, Alan Conrad},
  journal={IEEE Transactions on image processing},
  volume={21},
  number={12},
  pages={4695--4708},
  year={2012},
  publisher={IEEE}
}

@article{talebi2018nima,
  title={NIMA: Neural image assessment},
  author={Talebi, Hossein and Milanfar, Peyman},
  journal={IEEE transactions on image processing},
  volume={27},
  number={8},
  pages={3998--4011},
  year={2018},
  publisher={IEEE}
}

@inproceedings{zhang2025q,
  title={Q-eval-100k: Evaluating visual quality and alignment level for text-to-vision content},
  author={Zhang, Zicheng and Kou, Tengchuan and Wang, Shushi and Li, Chunyi and Sun, Wei and Wang, Wei and Li, Xiaoyu and Wang, Zongyu and Cao, Xuezhi and Min, Xiongkuo and others},
  booktitle={Proceedings of the Computer Vision and Pattern Recognition Conference},
  pages={10621--10631},
  year={2025}
}

@misc{novita,
  title        = {Novita AI},
  author       = {{Novita}},
  year         = {2024},
  howpublished = {\url{https://novita.ai}},
  note         = {Accessed: 2024-03-01}
}

@inproceedings{textssr,
  title={Textssr: Diffusion-based data synthesis for scene text recognition},
  author={Ye, Xingsong and Du, Yongkun and Tao, Yunbo and Chen, Zhineng},
  booktitle={Proceedings of the IEEE/CVF International Conference on Computer Vision},
  pages={17464--17473},
  year={2025}
}

@misc{pixart_alpha2023,
  title = {PixArt‑$\alpha$: Fast Training of Diffusion Transformer for Photorealistic Text-to-Image Synthesis},
  author = {Chen, Junsong and Yu, Jincheng and Ge, Chongjian and Yao, Lewei and Xie, Enze and Wu, Yue and Wang, Zhongdao and Kwok, James and Luo, Ping and Lu, Huchuan and Li, Zhenguo},
  year = {2023},
  url = {https://arxiv.org/abs/2310.00426},
  note = {arXiv preprint. Accessed: 2026-01-29}
}

@misc{sd35_large_turbo2024,
  title = {Stable Diffusion 3.5 Large Turbo},
  author = {Stability AI},
  year = {2024},
  url = {https://huggingface.co/stabilityai/stable-diffusion-3.5-large-turbo},
  note = {Model card / release. Accessed: 2026-01-29}
}

@misc{sd21_2022,
  title = {Stable Diffusion v2.1},
  author = {Stability AI},
  year = {2022},
  url = {https://huggingface.co/qualcomm/Stable-Diffusion-v2.1},
  note = {Model card. Accessed: 2026-01-29}
}

@misc{pixart_sigma2024,
  title = {PixArt‑$\sigma$: Weak-to-Strong Training of Diffusion Transformer for 4K Text-to-Image Generation},
  author = {Chen, Junsong and Ge, Chongjian and Xie, Enze and Wu, Yue and Yao, Lewei and Ren, Xiaozhe and Wang, Zhongdao and Luo, Ping and Lu, Huchuan and Li, Zhenguo},
  year = {2024},
  url = {https://arxiv.org/abs/2403.04692},
  note = {arXiv preprint / project page. Accessed: 2026-01-29}
}

@misc{sd35_medium2024,
  title = {Stable Diffusion 3 Medium},
  author = {Stability AI},
  year = {2024},
  url = {https://huggingface.co/stabilityai/stable-diffusion-3-medium},
  note = {Model card / release. Accessed: 2026-01-29}
}

@misc{kandinsky2_2023,
  title = {Kandinsky: An Improved Text-to-Image Synthesis with Image Prior and Latent Diffusion (Kandinsky 2)},
  author = {Razzhigaev, Anton and Shakhmatov, Arseniy and Maltseva, Anastasia and Arkhipkin, Vladimir and Pavlov, Igor and Ryabov, Ilya and Kuts, Angelina and Panchenko, Alexander and Kuznetsov, Andrey and Dimitrov, Denis},
  year = {2023},
  url = {https://arxiv.org/abs/2310.03502},
  note = {arXiv preprint. Accessed: 2026-01-29}
}

@misc{omnigen_2024,
  title = {OmniGen: Unified Image Generation},
  author = {Xiao, Shitao and Wang, Yueze and Zhou, Junjie and Yuan, Huaying and Xing, Xingrun and Yan, Ruiran and Li, Chaofan and Wang, Shuting and Huang, Tiejun and Liu, Zheng},
  year = {2024},
  url = {https://arxiv.org/abs/2409.11340},
  note = {arXiv preprint / project repo. Accessed: 2026-01-29}
}

@misc{sd3_medium2024,
  title = {Stable Diffusion 3 Medium (announcement)},
  author = {Stability AI},
  year = {2024},
  url = {https://stability.ai/news/stable-diffusion-3-medium},
  note = {Blog / release. Accessed: 2026-01-29}
}

@misc{sd35_large2024,
  title = {Stable Diffusion 3.5 Large},
  author = {Stability AI},
  year = {2024},
  url = {https://huggingface.co/stabilityai/stable-diffusion-3.5-large},
  note = {Model card. Accessed: 2026-01-29}
}

@misc{deepfloyd_if2022,
  title = {DeepFloyd IF (IF-I-M and related checkpoints)},
  author = {Stability AI},
  year = {2022},
  url = {https://stability.ai/news/deepfloyd-if-text-to-image-model},
  note = {Model card}
}

@misc{flux1_dev2026,
  title = {FLUX.1 [dev] (model card)},
  author = {Black Forest Labs (FLUX)},
  year = {2026},
  url = {https://huggingface.co/black-forest-labs/FLUX.1-dev},
  note = {Model card. Accessed: 2026-01-29}
}

@misc{cogview4_2025,
  title = {CogView4 (repo / model)},
  author = {Zai-Org},
  year = {2025},
  url = {https://github.com/zai-org/CogView4},
  note = {Project / model release. Accessed: 2026-01-29}
}

@misc{chatgpt_images2025,
  title = {ChatGPT Images (image generation) — OpenAI},
  author = {OpenAI},
  year = {2025},
  url = {https://help.openai.com/en/articles/8932459-creating-images-in-chatgpt},
  note = {Docs / feature page. Accessed: 2026-01-29}
}

@misc{flux11_pro2024,
  title = {FLUX1.1 Pro (product / model page)},
  author = {Black Forest Labs},
  year = {2024},
  url = {https://bfl.ai/models/flux-pro},
  note = {Vendor model page. Accessed: 2026-01-29}
}

@misc{flux2_max2025,
  title = {FLUX.2 [max] (model / product page)},
  author = {Black Forest Labs},
  year = {2025},
  url = {https://replicate.com/black-forest-labs/flux-2-max},
  note = {Model / API page. Accessed: 2026-01-29}
}

@misc{ideogram_v3_2025,
  title = {Ideogram 3 (Ideogram v3 / v3 turbo)},
  author = {Ideogram},
  year = {2025},    
  url = {https://ideogram.ai/features/3.0},
  note = {Product / model page. Accessed: 2026-01-29}
}

@misc{imagen4_fast2025,
  title = {Imagen 4 (Imagen 4 Fast) — Google / DeepMind},
  author = {Google DeepMind},
  year = {2025},
  url = {https://deepmind.google/models/imagen/},
  note = {Model page / announcement. Accessed: 2026-01-29}
}

@misc{nano_banana_pro2025,
  title = {Nano Banana Pro (Gemini 3 Pro Image)},
  author = {Google},
  year = {2025},
  url = {https://blog.google/innovation-and-ai/products/nano-banana-pro/},
  note = {Product blog. Accessed: 2026-01-29}
}

@misc{qwen_image2025,
  title = {Qwen-Image (model repo / release)},
  author = {Qwen Team / Alibaba},
  year = {2025},
  url = {https://github.com/QwenLM/Qwen-Image},
  note = {Model repo / release. Accessed: 2026-01-29}
}

@misc{sdxl2023,
  title = {SDXL: Improving Latent Diffusion Models for High-Fidelity Image Generation},
  author = {Podell, Dustin and English, Zion and Lacey, Kyle and Blattmann, Andreas and Dockhorn, Tim and Müller, Jonas and Penna, Joe and Rombach, Robin},
  year = {2023},
  url = {https://arxiv.org/abs/2307.01952},
  note = {arXiv preprint. Accessed: 2026-01-29}
}

@misc{seedream45_2025,
  title = {Seedream 4.5 (ByteDance / Seedream)},
  author = {ByteDance / Seedream},
  year = {2025},
  url = {https://byteplus.com/en/product/Seedream},
  note = {Product / API page. Accessed: 2026-01-29}
}

@misc{z_image_turbo2025,
  title = {Z-Image-Turbo (Tongyi-MAI / Alibaba)},
  author = {Tongyi-MAI},
  year = {2025},
  url = {https://huggingface.co/Tongyi-MAI/Z-Image-Turbo},
  note = {Model card / repo. Accessed: 2026-01-29}
}

@misc{easyocr,
  title        = {EasyOCR},
  author       = {{JaidedAI}},
  note = {Accessed: 2026-03-12},
  howpublished = {\url{https://github.com/JaidedAI/EasyOCR}},
}

@misc{rapidocr,
    title={{Rapid OCR}: OCR Toolbox},
    author={RapidAI Team},
    howpublished = {\url{https://github.com/RapidAI/RapidOCR}},
    year={2021}
}

@inproceedings{resnet,
  title={Deep residual learning for image recognition},
  author={He, Kaiming and Zhang, Xiangyu and Ren, Shaoqing and Sun, Jian},
  booktitle={Proceedings of the IEEE conference on computer vision and pattern recognition},
  pages={770--778},
  year={2016}
}

@article{vit,
  title={An image is worth 16x16 words: Transformers for image recognition at scale},
  author={Dosovitskiy, Alexey and Beyer, Lucas and Kolesnikov, Alexander and Weissenborn, Dirk and Zhai, Xiaohua and Unterthiner, Thomas and Dehghani, Mostafa and Minderer, Matthias and Heigold, Georg and Gelly, Sylvain and others},
  journal={arXiv preprint arXiv:2010.11929},
  year={2020}
}

@inproceedings{hyperiqa,
  title={Blindly assess image quality in the wild guided by a self-adaptive hyper network},
  author={Su, Shaolin and Yan, Qingsen and Zhu, Yu and Zhang, Cheng and Ge, Xin and Sun, Jinqiu and Zhang, Yanning},
  booktitle={Proceedings of the IEEE/CVF conference on computer vision and pattern recognition},
  pages={3667--3676},
  year={2020}
}

@inproceedings{chen2024drct,
  title={Drct: Diffusion reconstruction contrastive training towards universal detection of diffusion generated images},
  author={Chen, Baoying and Zeng, Jishen and Yang, Jianquan and Yang, Rui},
  booktitle={Forty-first International Conference on Machine Learning},
  year={2024}
}

@article{tuo2023anytext,
  title={Anytext: Multilingual visual text generation and editing},
  author={Tuo, Yuxiang and Xiang, Wangmeng and He, Jun-Yan and Geng, Yifeng and Xie, Xuansong},
  journal={arXiv preprint arXiv:2311.03054},
  year={2023}
}

\clearpage

\appendix


\phantomsection
\section*{Appendix Contents}
\addcontentsline{toc}{section}{Appendix Contents}

\begingroup
\setlength{\parindent}{0pt}
\setlength{\parskip}{2pt}

\newcommand{\apptocline}[3]{%
  \noindent\hspace*{#1}\hyperref[#2]{#3}\dotfill\pageref{#2}\par
}

\apptocline{0em}{app:impl}{Implementation Details}
\apptocline{1.5em}{app:impl:arch}{ANTIQA Architecture Specifications}
\apptocline{1.5em}{app:impl:training}{Training Recipe}
\apptocline{1.5em}{app:impl:proxy2mos}{OCR Confidence Mapping to MOS Scale}
\apptocline{1.5em}{app:impl:pooling}{Image-Level Aggregation Details and Pooling Ablations}
\apptocline{1.5em}{app:impl:speed}{Speed and Compute Measurement Protocol}
\apptocline{1.5em}{app:impl:vlm}{VLM Judging Protocol (Prompting and Score Parsing)}


\vspace{4pt}

\apptocline{0em}{app:results}{Extended Results}
\apptocline{1.5em}{app:results:appl_ocr_vlm}{Downstream task: predicting failures for OCR and VLM models}
\apptocline{1.5em}{app:iq_tq_decomposition}{Decomposing the correlation between overall quality (OQ-MOS) and text quality (TQ-MOS)}
\apptocline{1.5em}{app:results:pergen}{Per-Generator/Per-Prompt Breakdown on TIQA-Images}
\apptocline{1.5em}{app:results:ablations}{ANTIQA ablations}
\apptocline{1.5em}{app:results:vlm_analysis}{Analysis of VLM Behavior for Different Prompts}

\vspace{4pt}

\apptocline{0em}{app:data}{Extended Dataset Details}
\apptocline{1.5em}{app:data:inlab}{In-lab annotation markup}
\apptocline{1.5em}{app:data:postprocess}{Crops postprocessing}
\apptocline{1.5em}{app:data:prompts}{TIQA-Images Prompt List}
\apptocline{1.5em}{app:data:gens}{Generator List and Versioning}

\vspace{4pt}

\apptocline{0em}{app:human}{Subjective Study Protocol and Statistics}

\endgroup

\section{Implementation Details}
\label{app:impl}


\subsection{ANTIQA Architecture Specifications}
\label{app:impl:arch}
ANTIQA maps an input batch $I \in \mathbb{R}^{B\times 2\times H\times W}$ (grayscale concatenated with a Sobel edge map) to a scalar quality score per sample. The backbone is a 3-stage CNN with residual ConvB blocks, GroupNorm, SE gating, and strided downsampling, followed by per-scale adaptive pooling and an MLP regressor. Default hyperparameters are dropout $p{=}0.2$, pooling grid $G{=}2$, and SE reduction $r{=}16$.

\paragraph{Stem.} A single $\mathrm{Conv}_{3\times3}(2\!\to\!64)$ + GN + ReLU lifts the input to 64 channels at the original resolution. All convolutions in the network use $3{\times}3$ kernels with padding $1$ and no bias (GN absorbs it); GroupNorm uses $\min(8,C)$ groups, falling back to $1$ when $C$ is not divisible by $8$.

\paragraph{ConvB block.} Each ConvB is a residual block at constant channel count $C$ and resolution, with two convolutions, an inner ReLU, and channel-wise Dropout2d on the residual branch:
\[
\mathrm{ConvB}(x) = \mathrm{ReLU}\!\big(x + \mathrm{Drop2d}(\mathrm{GN}\circ\mathrm{Conv}\circ\mathrm{ReLU}\circ\mathrm{GN}\circ\mathrm{Conv})(x)\big).
\]
Dropout2d is applied at rate $0.5\,p$ in stages 1–2 and $p$ in stage 3.

\paragraph{DownScale.} Spatial downsampling and channel doubling are fused into a single strided convolution: \\ $\mathrm{DownScale}(x)=\mathrm{ReLU}(\mathrm{GN}(\mathrm{Conv}_{3\times3,\,s=2}(x)))$, mapping $C\!\to\!2C$ and halving the spatial dimensions.

\paragraph{Backbone stages.} Stage~1 stacks two ConvB(64) blocks, followed by DownScale to 128 channels at $H/2{\times}W/2$. Stage~2 stacks two ConvB(128) blocks, followed by DownScale to 256 channels at $H/4{\times}W/4$. Stage~3 stacks two ConvB(256) blocks. After each stage, an SE block~\cite{Hu_2018_CVPR} recalibrates channels: \\ $\mathrm{SE}(X)=X\odot\sigma(W_2\,\mathrm{ReLU}(W_1\,\mathrm{GAP}(X)))$, \\ with hidden width $\max(C/r,8)$.

\paragraph{Adaptive Pooling Block (APB).} At each scale $s\in\{0,1,2\}$ with $C_s\in\{64,128,256\}$, parallel adaptive average and max pooling reduce $X_s$ to a $G{\times}G$ grid; the flattened, concatenated descriptor $p_s\in\mathbb{R}^{2C_sG^2}$ is projected by a per-scale linear layer to $f_s=W_s p_s+b_s\in\mathbb{R}^{64}$.

\paragraph{Regression head.} The three per-scale descriptors are concatenated into $f=[f_0;f_1;f_2]\in\mathbb{R}^{192}$ and passed through an MLP of widths $192\!\to\!256\!\to\!128\!\to\!32\!\to\!1$ with ReLU activations and dropout (rates $p$ and $0.2\,p$), returning $\hat{y}\in\mathbb{R}$.

\paragraph{Resource profile.} ANTIQA contains 3.8M parameters and requires 31.5 GFLOPs per $256{\times}256$ crop.

\subsection{Training Recipe}
\label{app:impl:training}

We train the model in two stages: (i) pretraining on an OCR-confidence proxy target and
(ii) fine-tuning on human Mean Opinion Scores (MOS). In both stages, we minimize a weighted
combination of a regression loss and a ranking loss:
\begin{equation}
\mathcal{L} \;=\; \alpha \,\mathcal{L}_{\text{mse}} \;+\; (1-\alpha) \,\mathcal{L}_{\text{rank}},
\label{eq:loss_total}; \alpha=0.5.
\end{equation}

We optimize with AdamW using learning rate
$10^{-4}$, weight decay $w=\texttt{0.5}$,
batch size $B=4$, and train for
$E_{\text{pre}}=20$ epochs (pretraining) and
$E_{\text{ft}}=20$ epochs (fine-tuning).

We use a step schedule with step size $S=5$ epochs and decay factor $\gamma=0.5$.

All experiments are run with random seed $42$. Training is performed on NVIDIA A100 GPU with total compute of 35 GPU-hours.

\subsection{Baseline training configurations}
For comparison against ANTIQA, we train four baseline architectures on TIQA-Crops under matched protocols. All baselines follow the same two-stage curriculum as ANTIQA: $E_{\text{pre}}$ epochs of synthetic pretraining on the 110k OCR-confidence pseudo-labels, followed by $E_{\text{ft}}$ epochs of fine-tuning on the 10k human MOS labels. We optimize MSE with AdamW under a cosine schedule decaying to a minimum learning rate of $10^{-7}$, weight decay $10^{-4}$, and gradient clipping at $1.0$. Held-out validation splits of $500$ real and $1{,}000$ synthetic crops are used for model selection. All runs use a single GPU and the same random seed ($42$).

\paragraph{General-purpose backbones.}
ViT-Base/16~\cite{vit} and ResNet-50~\cite{resnet} are initialized from ImageNet-pretrained weights and equipped with a lightweight regression head (one hidden layer of width $512$, dropout $0.2$). Both receive crops resized to $224\times448$ to better match the typical aspect ratio of horizontal text regions; for ViT, we use the \texttt{vit\_base\_patch16\_224} variant with positional embeddings interpolated to the elongated input. Training uses bf16-mixed precision with $E_{\text{pre}}{=}4$ and $E_{\text{ft}}{=}10$ (cosine $T_{\max}{=}15$). ViT uses learning rates $5\times10^{-5}$ (synthetic) and $10^{-4}$ (real) at batch size $64$; ResNet-50 uses $10^{-4}$ and $3\times10^{-4}$ at batch size $128$.

\paragraph{No-reference IQA models.}
HyperIQA~\cite{hyperiqa} and TOPIQ-NR~\cite{topiq} are fine-tuned end-to-end from their official pretrained weights using fp16-mixed precision. HyperIQA is kept at its native $224\times224$ input, as its hypernetwork branch is tied to that resolution; we train it with $E_{\text{pre}}{=}4$ and $E_{\text{ft}}{=}2$, batch size $512$, and learning rates $10^{-5}$ (synthetic) and $2\times10^{-5}$ (real). TOPIQ-NR is fine-tuned at $224\times448$ with $E_{\text{pre}}{=}4$ and $E_{\text{ft}}{=}13$, batch size $128$, and learning rates $5\times10^{-6}$ and $10^{-5}$.

All four baselines are evaluated on the identical TIQA-Crops test split used for ANTIQA, ensuring that performance differences reflect architectural and training choices rather than data partitioning.

\subsection{OCR Confidence Mapping to MOS Scale}
\label{app:impl:proxy2mos}
\noindent We evaluated several candidate regression families to determine an appropriate parametric mapping from PaddleOCR confidence scores to subjective ratings (MOS). The candidates included random forest, a four-parameter logistic model, a five-parameter logistic model, support-vector regression with an RBF kernel, and ridge regression. Model selection showed that the five-parameter logistic provided the best fit to human judgments, in terms of correlations with MOS estimates on 10,000 crops from TIQA-Crops. Following this selection we fit the chosen parametric form within our neural optimal transport framework to obtain a final mapping from PaddleOCR confidence scores (interval $[0,1]$) to MOS (interval $[0,5]$).

\noindent The initial correlation between raw PaddleOCR confidence scores and MOS scores was: PLCC = $0.7734$. After mapping, PLCC improved slightly to PLCC = $0.8054$. Then, mapped scores for 110,000 crops were used at the pretrain stage of the proposed ANTIQA model.


\subsection{Image-Level Aggregation Details and Pooling Ablations}
\label{app:impl:pooling}
Given an image, we detect $N$ text crops. Each crop $i$ has a predicted quality score $s_i\in[0,5]$ and an area fraction $a_i\in(0,1]$ relative to the full image. Total text coverage is $A=\sum_{i=1}^N a_i$.

We define normalized area weights as $w_i(\alpha)=\frac{a_i^\alpha}{\sum_{j=1}^N a_j^\alpha}$ with $\alpha\ge 0$. Unless otherwise noted, methods below use $w_i(\alpha)$. Typical settings: $\alpha=1$ (simple area weighting), $\alpha=0.5$ (damped dominance), $\alpha=0$ (uniform).

\paragraph{List of pooling techniques.}
We map $\{(s_i,a_i)\}_{i=1}^N \mapsto S_{\mathrm{img}}$ using one of the following.

\begin{enumerate}
\item \textbf{Simple area-weighted mean.}
$S_{\mathrm{area}}=\sum_{i=1}^N w_i(1)\,s_i$.
(\emph{Parameter:} none; fixed $\alpha=1$.)
\item \textbf{Area$^\alpha$-weighted mean.}
$S_{\mathrm{mean}}(\alpha)=\sum_{i=1}^N w_i(\alpha)\,s_i$.
(\emph{Parameter:} $\alpha$; default $\alpha=0.5$.)
\item \textbf{Coverage-aware blend (prior + crop aggregate).}
$S_{\mathrm{cov}}=(1-\beta(A))\,s_0+\beta(A)\,S_{\mathrm{mean}}(\alpha)$ with $\beta(A)=1-e^{-A/A_0}$.
(\emph{Parameters:} prior $s_0$, coverage scale $A_0$, and $\alpha$; defaults $s_0\in\{10,5\}$, $A_0=0.03$, $\alpha=0.5$.)
\item \textbf{Softmin (log-sum-exp).} \\
$S_{\mathrm{softmin}}(\tau,\alpha)=-\tau\log\!\left(\sum_{i=1}^N w_i(\alpha)\,e^{-s_i/\tau}\right)$, $\tau>0$.
(\emph{Parameters:} $\tau$, $\alpha$; defaults $\tau=1.0$, $\alpha=0.5$.)
\item \textbf{Bottom-$k$ mean.}
Let $s_{(1)}\le \cdots \le s_{(N)}$ be sorted scores. Then $S_{\mathrm{bot}k}(k)=\frac{1}{k}\sum_{i=1}^k s_{(i)}$.
(\emph{Parameter:} $k$ or $k=\lceil \mathrm{frac}\cdot N\rceil$; defaults $\mathrm{frac}=0.2$ or $k\in\{1,2\}$ for small $N$.)
\item \textbf{Power mean (generalized mean).} \\
$S_{\mathrm{pm}}(p,\alpha)=\left(\sum_{i=1}^N w_i(\alpha)\,\tilde{s}_i^{\,p}\right)^{1/p}$ with $\tilde{s}_i=\max(s_i,\varepsilon)$.
(\emph{Parameters:} $p\neq 0$, $\varepsilon>0$, $\alpha$; defaults $p=-2$, $\varepsilon=10^{-3}$, $\alpha=0.5$.)
\end{enumerate}

If $N=0$, we either return the prior $s_0$ (when using the coverage-aware formulation) or mark the sample as ``no text detected'' and exclude it from text-quality evaluation, depending on protocol.

\begin{table*}[t]
\centering
\small
\setlength{\tabcolsep}{4pt}
\renewcommand{\arraystretch}{1.15}
\caption{Effect of aggregation on image-level correlation. Correlations were computed on the whole TIQA-Images dataset agains TQ-MOS (text-only quality score). Best score is bolded. Higher is better ($\uparrow$).}
\label{tab:agg_results}
\begin{tabular}{@{}l *{4}{S S} S S@{}}
\toprule
\multirow{2}{*}{Aggregation} &
\multicolumn{2}{c}{TOPIQ} &
\multicolumn{2}{c}{PaddleOCR} &
\multicolumn{2}{c}{Qwen3} &
\multicolumn{2}{c}{ANTIQA} &
\multicolumn{2}{c}{Average} \\
\cmidrule(lr){2-3}\cmidrule(lr){4-5}\cmidrule(lr){6-7}\cmidrule(lr){8-9}\cmidrule(lr){10-11}
& {PLCC$\uparrow$} & {SROCC$\uparrow$}
& {PLCC$\uparrow$} & {SROCC$\uparrow$}
& {PLCC$\uparrow$} & {SROCC$\uparrow$}
& {PLCC$\uparrow$} & {SROCC$\uparrow$}
& {PLCC$\uparrow$} & {SROCC$\uparrow$} \\
\midrule
Simple area-weighted mean          & 0.493 & 0.470 & 0.761 & 0.787 & 0.489 & 0.510 & 0.842 & 0.837 & 0.646 & 0.651 \\
Area$^\alpha$-weighted mean        & \textbf{0.512} & \textbf{0.483} & \textbf{0.780} & \textbf{0.792} & \textbf{0.503} & \textbf{0.527} & \textbf{0.844} & \textbf{0.841} & \textbf{0.660} & \textbf{0.661} \\
Coverage-aware blend               & 0.241 & 0.204 & 0.476 & 0.455 & 0.213 & 0.221 & 0.511 & 0.517 & 0.360 & 0.349 \\
Softmin (log-sum-exp)              & 0.485 & 0.466 & 0.749 & 0.771 & 0.463 & 0.490 & 0.819 & 0.813 & 0.629 & 0.635 \\
Bottom-$k$ mean                    & 0.461 & 0.459 & 0.712 & 0.756 & 0.448 & 0.475 & 0.793 & 0.802 & 0.604 & 0.623 \\
Power mean                         & 0.479 & 0.460 & 0.738 & 0.765 & 0.451 & 0.477 & 0.806 & 0.802 & 0.619 & 0.626 \\
\bottomrule
\end{tabular}
\end{table*}

Table~\ref{tab:agg_results} shows that area($^\alpha$)-weighted mean pooling is the clear winner across all four OCR/TIQA pipelines, achieving the best average correlation with TQ-MOS (PLCC 0.660 / SROCC 0.661) and consistently topping each individual model (e.g., PaddleOCR 0.780/0.792, ANTIQA 0.844/0.841). Relative to simple area-weighting, the gains are modest but systematic, suggesting that damping the dominance of very large crops (via ($\alpha \le 1$)) better matches human text-quality judgments by letting multiple regions contribute. In contrast, methods that explicitly emphasize worst-case regions (softmin, bottom-(k), power mean with negative (p)) generally reduce correlation, indicating that penalizing a few low-quality crops overstates their impact at the image level. The coverage-aware blend collapses to much lower correlations, implying that injecting a global prior based on total text coverage is misaligned with a *text-only* MOS target (and likely dilutes signal when text is present). Overall, simple weighted averaging is robust, but area($^\alpha$) pooling provides the most reliable improvement, making it the best default aggregator.

\subsection{Speed and Compute Measurement Protocol}
\label{app:impl:speed}
The FPS rate for the ANTIQA (proposed), PaddleOCR, RapidOCR, EasyOCR, SAR,  HyperIQA and TOPIQ models was measured by running each model 500 times on the same crop with a resolution of \(256\times256\) followed by taking the minimum time. The execution time of Qwen3 and GLM 4.6 was calculated based on 50 requests to the service API novita.ai \cite{novita} with the prompt to evaluate the same \(256\times256\) crop. However, the measured time includes not only the model calculations themselves, but also additional operations, we consider them negligible.

FLOPs for ANTIQA were estimated per-forward using off-the-shelf profiler on the same input crop (batch size 1, same \(256\times256\) resolution.

\subsection{VLM Judging Protocol (Prompting and Score Parsing)}
\label{app:impl:vlm}
Qwen3 \cite{qwen3} and GLM 4.6 \cite{glm46} were accessed via the API of the service novita.ai \cite{novita}. All the crops from TIQA-Crops and images from TIQA-Images were uploaded to the Internet in order to enable VLM to view and download them. Prompts are presented below:

\begin{figure*}[h!]
\begin{tcolorbox}[promptbox]
\textbf{Prompt for TIQA-Crops test crops:} \texttt{"Imagine that you are an OCR model and that you are looking at the quality of the text in the picture. Your task is to produce a real number from 0 to 5, which will represent the quality of the text in the picture in terms of artifacts. In real OCR models, this number is called ocr\_score or confidence\_score. You can use fractional values like 4.33 or 1.23."}
\end{tcolorbox}
\end{figure*}
\begin{figure*}[h!]
\begin{tcolorbox}[promptbox]
\textbf{Prompt for TIQA-Images:} 
\texttt{"You will see images generated by artificial intelligence, which often look high-quality, but may contain subtle artifacts in the text. Your task is to evaluate ONLY the quality of the text in the image. \\ What NOT to consider: any content of image other then text, semantic correctness of the text, realistic and naturalness, sharpness, noise, and other processing defects.  \\ Take into account ONLY visible defects in the text, non-existent letters, and text distortions, as they affect the overall impression. \\ Score 0: Very low quality. \\  Score 1: Strong distortion of the text throughout the image, a lot of interference. \\  Score 2: Text distortion is very noticeable. \\   Score 3: Some defects in the text are noticeable on closer inspection. \\  Score 4: Almost all the text looks perfect, there are minor defects. \\  Score 5: The image does not contain any text defects, the whole text is perfectly readable. \\ \\  You can use fractional values such as 4.33 or 1.23. In your answer, give only one real number."}
\end{tcolorbox} \end{figure*}

Settings such as temperature and max\_tokens, were not changed and those provided by the novita.ai \cite{novita} were used, and the values were set to temperature = 1.0, max\_tokens = the size of the context window.

\section{Extended Results}
\label{app:results}

\subsection{Downstream task: predicting failures for OCR and VLM models}
\label{app:results:appl_ocr_vlm}

Another downstream task for which potential TIQA models are suitable is the prediction of recognition errors and hallucinations produced by OCR and VLM systems. To demonstrate this, we ran a controlled experiment and analysis pipeline.
\paragraph{Overview.} We first collected clean text crops from generated images, then used OCR and VLMs to obtain reliable baseline recognized texts, discarding low-confidence samples. Each crop was then progressively degraded using a six-stage distortion pipeline. OCR, VLM, ANTIQA, and TOPIQ were applied to all distorted versions in order to obtain texts, confidences and quality predictions. Finally, we measured recognition errors using normalized Levenshtein similarity and analyzed how these errors correlate with OCR/VLM confidence scores and ANTIQA/TOPIQ predictions. High correlation indicates that TIQA scores effectively capture image and text degradation relevant to OCR and VLM hallucinations.

\medskip
We denote the \(k\)-th corrupted version of crop \(n\) by \(c_{n,k}\) (with \(k=1,\dots,6\)). Let \(\mathrm{ref}_n\) be the ground-truth text (from the clean crop). Denote OCR/VLM recognized text on \(c_{n,k}\) by \(\mathrm{hyp}_{n,k}\). We define the normalized Levenshtein error
\[
\mathrm{nsim}_{n,k}
\;=\;
1 - \frac{d_{\mathrm{lev}}(\mathrm{ref}_n,\mathrm{hyp}_{n,k})}
{\max\{\lvert\mathrm{ref}_n\rvert,\lvert\mathrm{hyp}_{n,k}\rvert,1\}},
\]
where \(d_{\mathrm{lev}}(\cdot,\cdot)\) is the Levenshtein edit distance; \(\mathrm{nsim}_{n,k}\in[0,1]\) with \(1\) meaning perfect match.

\textbf{Distortion of crops.}
\label{par:distortion-crops}
A modified TextSSR \cite{textssr} pipeline was used to generate distorted crops. Initially, the TextSSR \cite{textssr} pipeline uses areas of text cut out of images as conditioning, as well as rendered glyphs on a white background.

\begin{figure}[h!]
    \centering
\includegraphics[width=\linewidth]{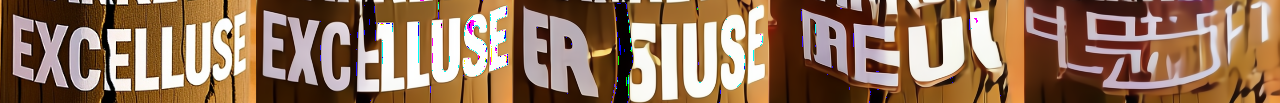}
    \caption{An example of gradual crop distortion, from left to right.}
    \label{fig:crop_distortion}
\end{figure}

We took these rendered glyphs at inference time and distorted them in two ways: by applying Gaussian blur with different radii and by applying JPEG compression with different quality settings. Intuitively, this can be described as “blurring the eyes” of the diffusion model, which leads to poorer-quality text generation. We also ran the TextSSR pipeline twice and three times in succession, replacing the original clean crops with the generated ones and thereby degrading the input data for the next iteration. An example of gradual distortion in a Figure~\ref{fig:crop_distortion}.

\begin{figure*}[h!] \centering \includegraphics[width=\linewidth]{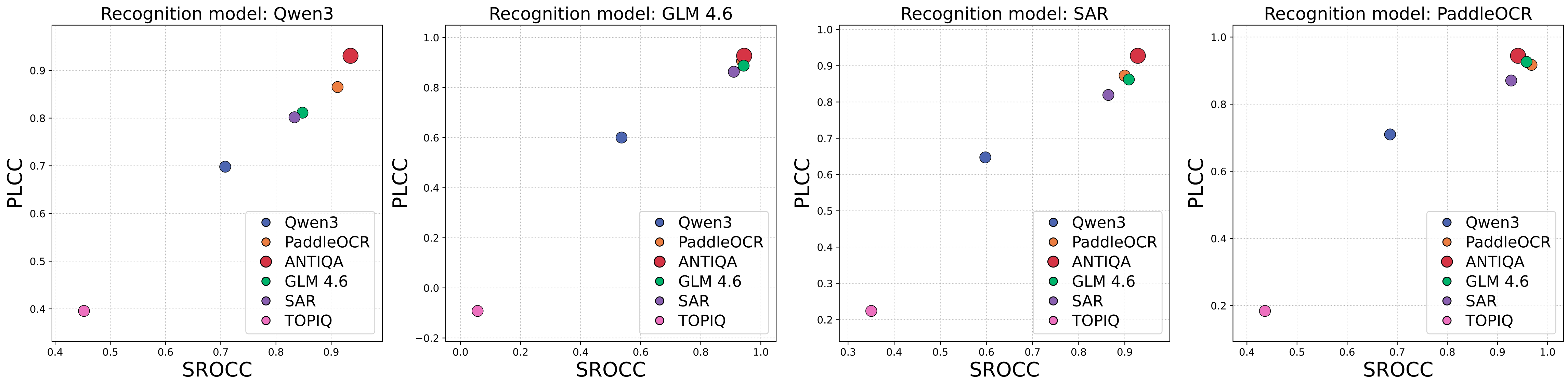} \caption{Binned mean normalized Levenshtein similarity as a function of predicted by TIQA models score.}\label{fig:scatter_for_textssr}
\end{figure*}

\textbf{Results.} Figure~\ref{fig:scatter_for_textssr} shows PLCC and SROCC correlations for all tested models. The proposed ANTIQA model is predominantly ahead, especially for the task of detecting Qwen3 hallucination (the left figure). PaddleOCR and GLM 4.6 are not far behind, which also perform well, especially in the task of detecting PaddleOCR errors (fourth graph) and detecting GLM 4.6 hallucinations (second graph).


\subsection{ANTIQA as a Feature Extractor for AI-Generated Image Detection}
\label{app:results:antiqa_for_aigi}

We investigate whether the internal representations of ANTIQA carry signal useful for a different downstream task: distinguishing real pictures from AI-generated ones. The intuition is that text rendered by current text-to-image generators is one of the most fragile aspects of synthetic content, so a model trained to score text quality should implicitly capture cues that betray a generated image. We further study whether such cues are complementary to those of a dedicated deepfake detector.

\paragraph{Data.}
We start from a $70{,}000$  subset of real text-containing images sampled from AnyWord~\cite{tuo2023anytext}, which aggregates images with visible text from sources such as LAION~\cite{schuhmann2022laion5b}. For each image, AnyWord provides a caption together with the literal text appearing in the image, which we turn into prompts of the form \texttt{"\{caption\}. Text in image: \{visible text\}"} to synthesize generated counterparts. We use $30$ generators in total, partitioned into \textbf{Pool~A} ($20$ generators) and \textbf{Pool~B} ($10$ generators), and evaluate all models on the full Pool~A$\cup$Pool~B test set. Each split is balanced with an equal number of real source images, and the task is framed as binary classification with labels $y\in\{0,1\}$ (real vs.\ generated). For \textbf{Pool~A}, we generated $40$k images for training ($2{,}000$ per generator) and $20$k images for testing ($1{,}000$ per generator). For \textbf{Pool~B}, we generated $10$k images, used exclusively for testing ($1{,}000$ per generator), yielding a total of $70{,}000$ generated images paired with the $70{,}000$ real images sampled from AnyWord.

\paragraph{ANTIQA features and fusion.}
Each image is first decomposed into text crops using the same detection and rectification pipeline as in TIQA-Crops. Every crop is then passed through ANTIQA, from which we extract the most informative representation: the $192$-dimensional fused multi-scale descriptor produced by the APB block before the regression head. These per-crop descriptors form a variable-length set per image, which we aggregate into a fixed-size representation through a simple Mean adapter: per-crop features are pooled by mean, concatenated with the global DRCT image embedding, and passed through an MLP head trained with MSE loss to predict the binary label.

\paragraph{Training budgets.}
The two models compared in this study are trained on the same underlying Pool~A training set. The standalone DRCT~\cite{chen2024drct} detector is fine-tuned on the full Pool~A training split of $40{,}000$ generated images ($2{,}000$ per generator) together with an equal number of real images. The \emph{Mean adapter}, in contrast, partitions this same $40$k split into two disjoint subsets: its underlying DRCT detector is fine-tuned on $32{,}000$ generated images ($1{,}600$ per generator) plus an equal number of reals, and its fusion MLP is subsequently trained on the remaining $8{,}000$ generated images ($400$ per generator) plus an equal number of reals. Both models are evaluated on the same test sets: the Pool~A test split ($20{,}000$ images) and the held-out Pool~B test split ($10{,}000$ images).

\paragraph{Results and discussion.}
Averaged over $k=3$ runs with different seeds, the standalone DRCT detector reaches a ROC-AUC of $0.967$ on the combined Pool~A$\cup$Pool~B test set, while the Mean adapter that fuses DRCT with ANTIQA features improves this to $0.973$, even though its underlying detector sees only $32$k of the $40$k training images available to the standalone baseline. This gain, modest in absolute terms but meaningful at the upper end of the ROC-AUC scale, supports the interpretation that the representations learned for text-quality assessment capture cues at least partially orthogonal to those of a dedicated deepfake detector.

\subsection{Decomposing the correlation between overall quality (OQ-MOS) and text quality (TQ-MOS)}
\label{app:iq_tq_decomposition}

\begin{table}[h]
\centering
\small
\setlength{\tabcolsep}{6pt}
\caption{Decomposing the correlation between human overall quality (OQ-MOS) and text quality (TQ-MOS) on TIQA-Images.
TIQA-Images contains $P{=}30$ prompts, $G{=}10$ generators, and $K{=}5$ seeds per (prompt, generator) pair. 
}
\begin{tabular}{lccc}
\toprule
\textbf{Correlation level} & \textbf{\# points} & \textbf{SROCC} \\
\midrule
Pooled (all images) & $GPK=1500$ & 0.78\\
Between generators (means) & $G=10$ & 0.98 \\
Within (prompt, generator) & $S=300$ & 0.51 $\pm$ 0.43 \\
\quad\quad (median) & $S=300$ & 0.59 \\
\bottomrule
\end{tabular}
\label{tab:iq_tq_decomposition}
\end{table}

TIQA-Images contains generations from $G=10$ text-to-image generators evaluated on $P=30$ prompts, with $K=5$ random seeds per (generator, prompt) pair, for a total of $GPK=1500$ images. Each image $(g,p,k)$ has two human mean-opinion scores (MOS): overall image quality $OQ_{g,p,k}$ (OQ-MOS) and text rendering quality $TQ_{g,p,k}$ (TQ-MOS). We report Pearson linear correlation (PLCC) and Spearman rank correlation (SROCC) as measures of association.
A pooled correlation computed over all images can be inflated by between-generator differences. If some generators are systematically better at both rendering text and producing overall high-quality images, the pooled correlation may appear large even if, within a fixed generator and prompt, seed-to-seed variation in text quality is unrelated to seed-to-seed variation in overall quality. To address this concern, we compute correlations at multiple levels of control (Table~\ref{tab:iq_tq_decomposition}).

We first compute the association between overall quality and text quality over all images:
\[
\rho_{\mathrm{all}} = \mathrm{corr}\big(\{OQ_{g,p,k}\}, \{TQ_{g,p,k}\}\big),
\]
where $\mathrm{corr}(\cdot,\cdot)$ is either PLCC or SROCC and the sets range over all $g \in \{1,\dots,G\}$, $p \in \{1,\dots,P\}$, and $k \in \{1,\dots,K\}$. This yields the pooled OQ-MOS--TQ-MOS correlation reported in the main text (SROCC $\approx 0.78$).

To isolate how much of the association is explained by systematic differences between generators, we average scores within each generator across all prompts and seeds:
\[
\overline{OQ}_g = \frac{1}{PK}\sum_{p=1}^{P}\sum_{k=1}^{K} OQ_{g,p,k},
\qquad
\overline{TQ}_g = \frac{1}{PK}\sum_{p=1}^{P}\sum_{k=1}^{K} TQ_{g,p,k}.
\]
We then compute the correlation across the $G=10$ generator-level points:
\[
\rho_{\mathrm{gen}} = \mathrm{corr}\big(\{\overline{OQ}_g\}_{g=1}^{G}, \{\overline{TQ}_g\}_{g=1}^{G}\big).
\]
Empirically, this between-generator association is extremely high (PLCC=0.96, SROCC=0.98), indicating that generators that render text better are almost always judged better overall on these text-heavy prompts.

To test whether the association persists when generator and prompt are held fixed, we compute a within-pair Spearman correlation across the $K=5$ seeded samples for each (generator, prompt) pair:
\[
\rho_{g,p} = \mathrm{SROCC}\Big(\{OQ_{g,p,k}\}_{k=1}^{K}, \{TQ_{g,p,k}\}_{k=1}^{K}\Big).
\]
This produces $S=GP=300$ within-pair correlations. We summarize the distribution of $\rho_{g,p}$ by reporting its mean, standard deviation, and median across the 300 pairs. Empirically, we observe a strongly positive within-pair association (mean SROCC=0.51, std=0.43, median=0.59), showing that even for the same generator on the same prompt, seeds that yield better text are typically also rated better overall.
Overall, the near-perfect between-generator correlation shows that text quality is a major axis separating systems on text-heavy prompts, while the strong within-(generator, prompt) correlations show that the association is not merely an artifact of comparing different generators. Instead, seed-level improvements in text rendering quality tend to coincide with seed-level improvements in overall perceived quality in this regime (Table~\ref{tab:iq_tq_decomposition}).

\subsection{Per-Generator/Per-Prompt Breakdown on TIQA-Images}
\label{app:results:pergen}

\begin{figure*}[h!]
    \centering
\includegraphics[width=0.93\linewidth]{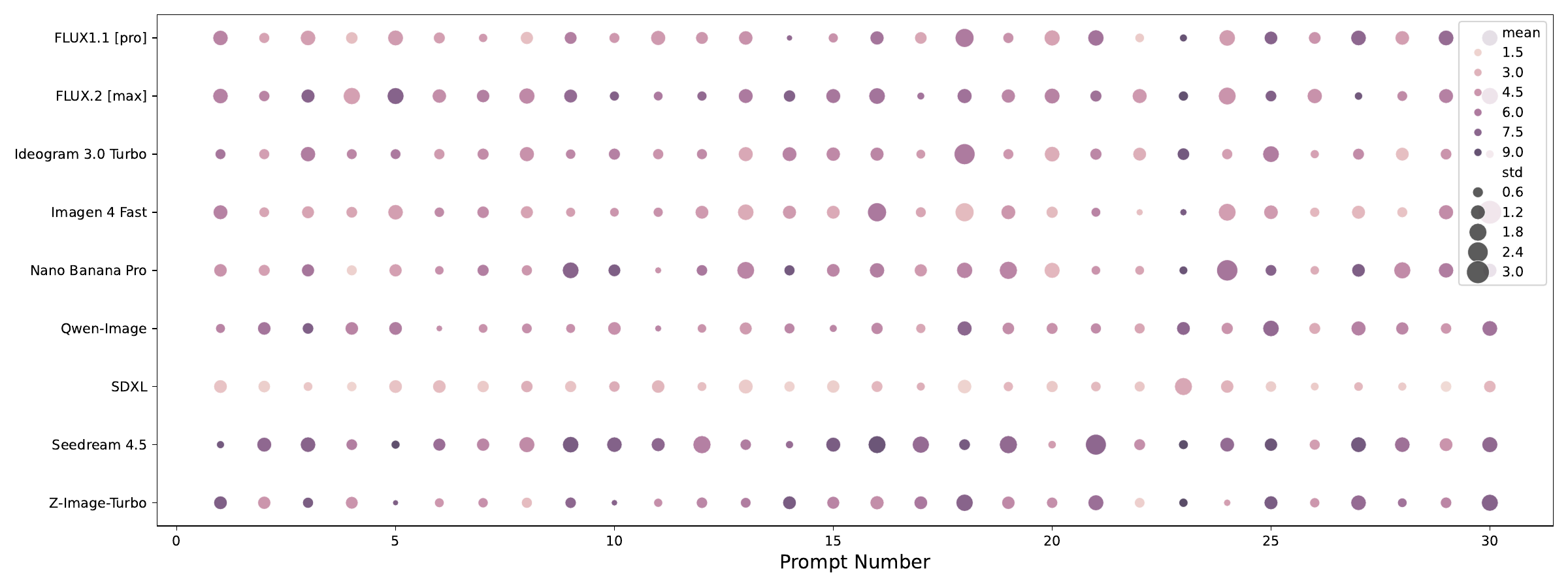}
    \caption{Prompt and seed dependencies across text-to-image models ON TIQA-Images. Colour encodes the mean TQ-MOS per prompt, and marker size encodes the standard deviation across five seed generations, capturing within-prompt sampling variability.}
    \label{fig:breakdown}
\end{figure*}

The Figure~\ref{fig:breakdown} decomposes performance into expected accuracy and sampling reliability by plotting the mean score (colour) and the standard deviation across five seed generations (marker size) for each prompt. Across prompts, several models attain similarly strong means, but their seed-level dispersion differs markedly: Seedream 4.5 and Z-Image-Turbo frequently combine high averages with larger variability, while SDXL is consistently lower on average yet comparatively tight. This gap matters for real use, since a high mean with high seed variance implies a non-trivial chance of poor single-shot outputs and a greater need for resampling. These results motivate reporting seed dispersion alongside mean, and adopting risk-aware summaries such as lower quantiles or pass rates at a fixed quality threshold to better reflect deployment-facing robustness.

\subsection{ANTIQA ablations}
\label{app:results:ablations}
Table~\ref{tab:antiqa_ablations} confirms that ANTIQA’s main components materially affect performance on TIQA-Crops. The full model reaches 0.942 PLCC / 0.935 SROCC, while removing OCR-confidence pretraining causes a clear drop to 0.887 / 0.881, highlighting the value of leveraging the 110k proxy-labeled crops before finetuning on the 10k MOS set. Architecturally, strip convolutions are crucial: replacing the proposed strip conv blocks with standard convolutions substantially degrades performance to 0.890 / 0.876, consistent with the need for directionally-aware modeling of text-like structures. Removing multi-scale feature extraction yields the largest drop (0.834 / 0.836), further indicating that text artifacts must be captured across scales.
\begin{table}[h]
\centering
\small
\caption{ANTIQA ablations on TIQA-Crops. Full-model numbers are from Table 1 in the paper; the provided PDF does not include the ablation-result values.}
\begin{tabular}{lcc}
\toprule
\textbf{Variant} & \textbf{PLCC} & \textbf{SROCC} \\
\midrule
Full ANTIQA (proposed) & 0.942 & 0.935 \\
\midrule
\makecell[l]{w/o OCR-confidence \textbf{pretraining}} & 0.887 & 0.881 \\
\makecell[l]{w/o \textbf{neural optimal transport} mapping\\(use raw OCR confidence)} & 0.933 & 0.927 \\
\midrule
w/o \textbf{Sobel edge-map} input (grayscale only) & 0.920 & 0.923 \\
w/o \textbf{strip conv} blocks (use standard conv) & 0.890 & 0.876 \\
w/o \textbf{SE} channel gating & 0.905 & 0.908 \\
w/o \textbf{multi-scale} feature extraction & 0.834 & 0.836 \\
w/o \textbf{avg+max} in APB (avg-only) & 0.895 & 0.901 \\
w/o \textbf{avg+max} in APB (max-only) & 0.903 & 0.904 \\
\bottomrule
\end{tabular}
\label{tab:antiqa_ablations}
\end{table}

\subsection{Analysis of VLM Behavior for Different Prompts}
\label{app:results:vlm_analysis}
We design a controlled prompt-sensitivity study using four variants of the same base prompt, each adding a progressively larger amount of task-specific detail. For evaluation, we sample 60 images from TIQA-Images at random and run Qwen3 three independent times per prompt–image pair to account for generation stochasticity. Results show that performance varies noticeably across prompt variants, indicating that Qwen3 is highly prompt-dependent: changes in wording and the level of instruction detail can lead to non-trivial shifts in correlation with MOS. This further underscores the unreliability of current VLM-based evaluation: despite identical inputs, performance can fluctuate substantially with prompt edits and across repeated runs, indicating limited robustness and weak reproducibility.

\begin{table}[h]
\centering
\caption{Correlation of Qwen3 with MOS scores using different prompts on 60 images.}
\begin{tabular}{lcccc}
\toprule
& \multicolumn{2}{c}{TQ--MOS} & \multicolumn{2}{c}{OQ--MOS} \\
\cmidrule(lr){2-3}\cmidrule(lr){4-5}
Prompt & PLCC & SROCC & PLCC & SROCC \\
\midrule
1 & 0.6785 & 0.6912 & 0.6628 & 0.6866 \\
2 & 0.4342 & 0.4983 & 0.4772 & 0.5392 \\
3 & 0.6749 & 0.6505 & 0.7043 & 0.7066 \\
4 & 0.5683 & 0.6004 & 0.5706 & 0.5708 \\
\bottomrule
\end{tabular}
\end{table}

\section{Extended Dataset Details}
\label{app:data}

\subsection{In-lab detector annotation markup}
\label{app:data:inlab}

Before committing to PP-OCRv5~\cite{paddle} as the text-region detector for the TIQA pipeline, we conducted a small in-lab study to quantify how reliably different open-source detectors localize potentially artifact-prone text regions in generator outputs. The goal was not to measure raw detection quality in the classical sense (bounding-box overlap against ground truth) but rather \emph{recall of candidate regions} from a human perspective: for a downstream text-quality annotation task, missing a distorted text region is far worse than producing a spurious detection, since the latter can be filtered out at the annotation stage while the former silently removes data from the study.

\paragraph{Detectors.}
We compared three popular open-source text detectors: PP-OCRv5 (the detection module of PaddleOCR)~\cite{paddle}, EasyOCR~\cite{easyocr}, and RapidOCR~\cite{rapidocr}. Each detector was used with its default configuration and no post-processing beyond its built-in thresholds, so the comparison reflects what a practitioner would obtain out of the box.

\paragraph{Data and protocol.}
We sampled a balanced subset of the images from which TIQA-Crops was built, taking $100$ images per generator, for a total of $1{,}200$ images spanning the 12 generators listed in Table~\ref{tab:generators}. Every image was processed independently by each of the three detectors, and the resulting candidate text regions were visualized as red bounding boxes overlaid on the original image.

We then built a minimal annotation interface that showed a lab annotator one such visualization at a time, together with two buttons labeled $0$ and $1$. The annotator was instructed to press $0$ if the detector had found \emph{all} regions they considered to potentially contain text, and $1$ if the detector had missed \emph{at least one} such region. Crucially, false positives---bounding boxes placed on regions that obviously contain no text at all---were explicitly excluded from the criterion, because such spurious detections are discarded later during the crop-level annotation stage and therefore do not hurt the final dataset. The metric of interest is thus the fraction of images on which the detector's recall of human-perceived text regions was complete; we refer to it as the \emph{no-miss rate}.

\paragraph{Results.}
On the $1{,}200$-image evaluation set, PP-OCRv5 consistently produced the most complete set of detections, achieving a no-miss rate of $98\%$, compared to $94\%$ for EasyOCR and $89\%$ for RapidOCR.

\subsection{Crops postprocessing}
\label{app:data:postprocess}
Although PP-OCRv5~\cite{paddle} provides strong recall of candidate text regions, its detection module occasionally produces false positives in areas that contain no textual content. To mitigate the contamination of the dataset with such spurious crops, we introduced a two-stage filtering procedure at the preprocessing level. First, the recognition module of PP-OCRv5 was applied to every detected region, and crops whose recognition confidence fell below a threshold of $0.2$ were discarded. Second, we removed all crops with a height of less than $20$ pixels, as these were empirically found to lack sufficient resolution for reliable quality assessment. Each retained crop was subsequently rectified to a canonical horizontal orientation via a perspective transform. Finally, as a post-annotation filtering step, crops that received a subjective score of $0$ were excluded from the dataset, since such ratings indicate the absence of any discernible textual content and therefore contribute no informative signal to the text-quality assessment task.

\subsection{TIQA-Images Prompt List (Text-Heavy Prompts)}
\label{app:data:prompts}
Here we show three examples out of 30 prompts created for TIQA-Images dataset:
\begin{tcolorbox}[promptbox]
\textbf{Prompt 1.}
A photorealistic airport departures board, wide angle. The board has 30+ rows with small text, columns, and mixed alphanumeric codes. Include slight motion blur from people passing, but the board itself should remain readable. \\ Columns:\\Time | Flight | Destination | Gate | Status \\ Include these exact rows somewhere:\\ 06:45  KL 1023  AMSTERDAM   D12  BOARDING\\ 09:10  QR 274   DOHA        B07  ON TIME\\ 12:30  TK 1952  ISTANBUL    C03  LAST CALL\\ 18:05  EK 148   DUBAI       A19  DELAYED 45 MIN\\Add a ticker line at bottom in tiny text:\\Security notice: do not leave baggage unattended.
\end{tcolorbox}

\begin{tcolorbox}[promptbox]
\textbf{Prompt 2.}
A TV karaoke screen in a dim room. The screen shows large lyrics and a second line of smaller subtitles, plus a dense list of upcoming lines in tiny text on the side. Add mild moiré from filming a screen.\\Lyrics (large):\\WE KEEP RUNNING THROUGH THE NIGHT\\UNTIL THE MORNING LIGHT\\Small subtitle line:\\(sing along)\\Include a tiny song code:\\Track ID: K-10492
\end{tcolorbox}

\begin{tcolorbox}[promptbox]
\textbf{Prompt 3.}
A clean comic page with multiple panels. Each panel has speech bubbles with readable text, plus tiny sound effects and small panel captions. Some speech bubbles are curved.\\Include these exact bubble lines:\\We can’t trust the first render.\\Zoom in—look at the kerning!\\Why is every third letter wrong?\\Add tiny sound effect text:\\*buzz*  *click*
\end{tcolorbox}
Full list of prompts will be released alongside the dataset.

\subsection{Generator List and Versioning}
\label{app:data:gens}

The models used to generate images for both datasets are shown in the Table~\ref{tab:generators}. All settings and initial parameters were set to default.

\begin{table}[ht]
  \centering
  \caption{Generators used for TIQA datasets}
  \label{tab:generators}
  \begin{tabular}{c  l  l}
    \toprule
    \# & \textbf{TIQA-Crops} & \textbf{TIQA-Images} \\
    \midrule
    1  & PixArt Alpha \cite{pixart_alpha2023}            & GPT Image 1.5 \cite{chatgpt_images2025} \\
    2  & SD 3.5 Large Turbo \cite{sd35_large_turbo2024} & FLUX1.1 [pro] \cite{flux11_pro2024} \\
    3  & SD 2.1 \cite{sd21_2022}                         & FLUX.2 [max] \cite{flux2_max2025} \\
    4  & PixArt Sigma \cite{pixart_sigma2024}          & Seedream 4.5 \cite{seedream45_2025} \\
    5  & SD 3.5 Medium \cite{sd35_medium2024}           & Ideogram 3.0 Turbo \cite{ideogram_v3_2025} \\
    6  & Kandinsky 2 \cite{kandinsky2_2023}             & Imagen 4 Fast \cite{imagen4_fast2025} \\
    7  & Omnigen \cite{omnigen_2024}                     & Z-Image-Turbo \cite{z_image_turbo2025} \\
    8  & SD 3 Medium \cite{sd3_medium2024}              & Nano Banana Pro \cite{nano_banana_pro2025} \\
    9  & SD 3.5 Large \cite{sd35_large2024}             & Qwen-Image \cite{qwen_image2025} \\
    10 & DeepFloyd IF \cite{deepfloyd_if2022}           & SDXL \cite{sdxl2023} \\
    11 & FLUX.1 Dev \cite{flux1_dev2026}                \\
    12 & CogView4 \cite{cogview4_2025}                \\
    \bottomrule
  \end{tabular}
\end{table}

\subsection{Examples}
\label{app:data:examples}

\begin{figure}[h!] \centering \includegraphics[width=1.0\linewidth]{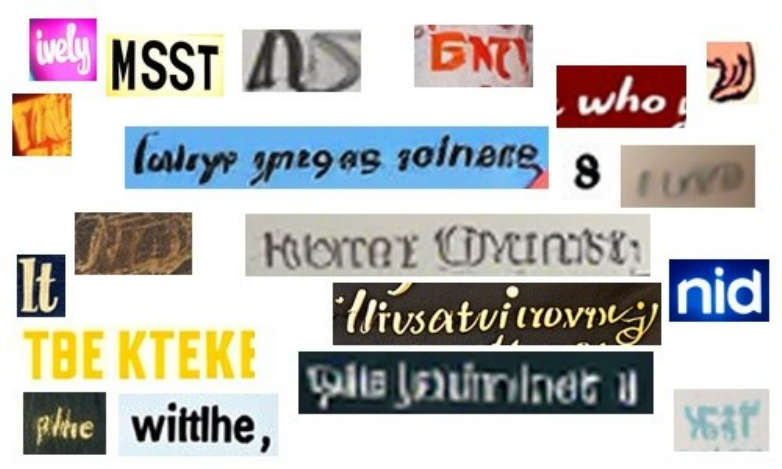} \caption{TIQA-Crops examples}\label{fig:random_grid}
\end{figure}

\begin{figure*}[h!] 
\centering 
\includegraphics[width=\linewidth]{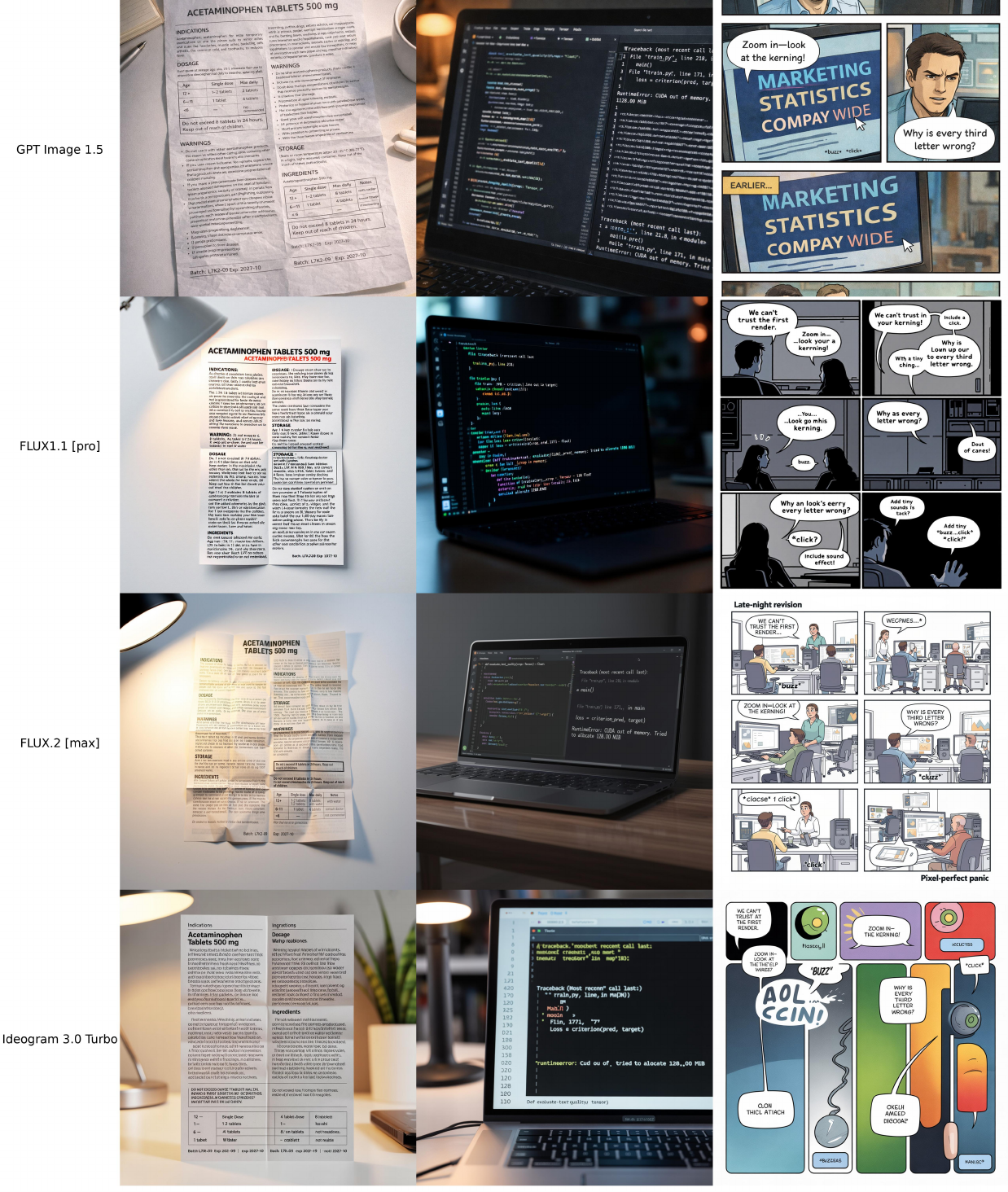} \caption{TIQA-Images examples for overall quality}
\label{fig:images_examples}
\Description{TODO}
\end{figure*}

\begin{figure*}[h!] \centering \includegraphics[width=0.9\linewidth]{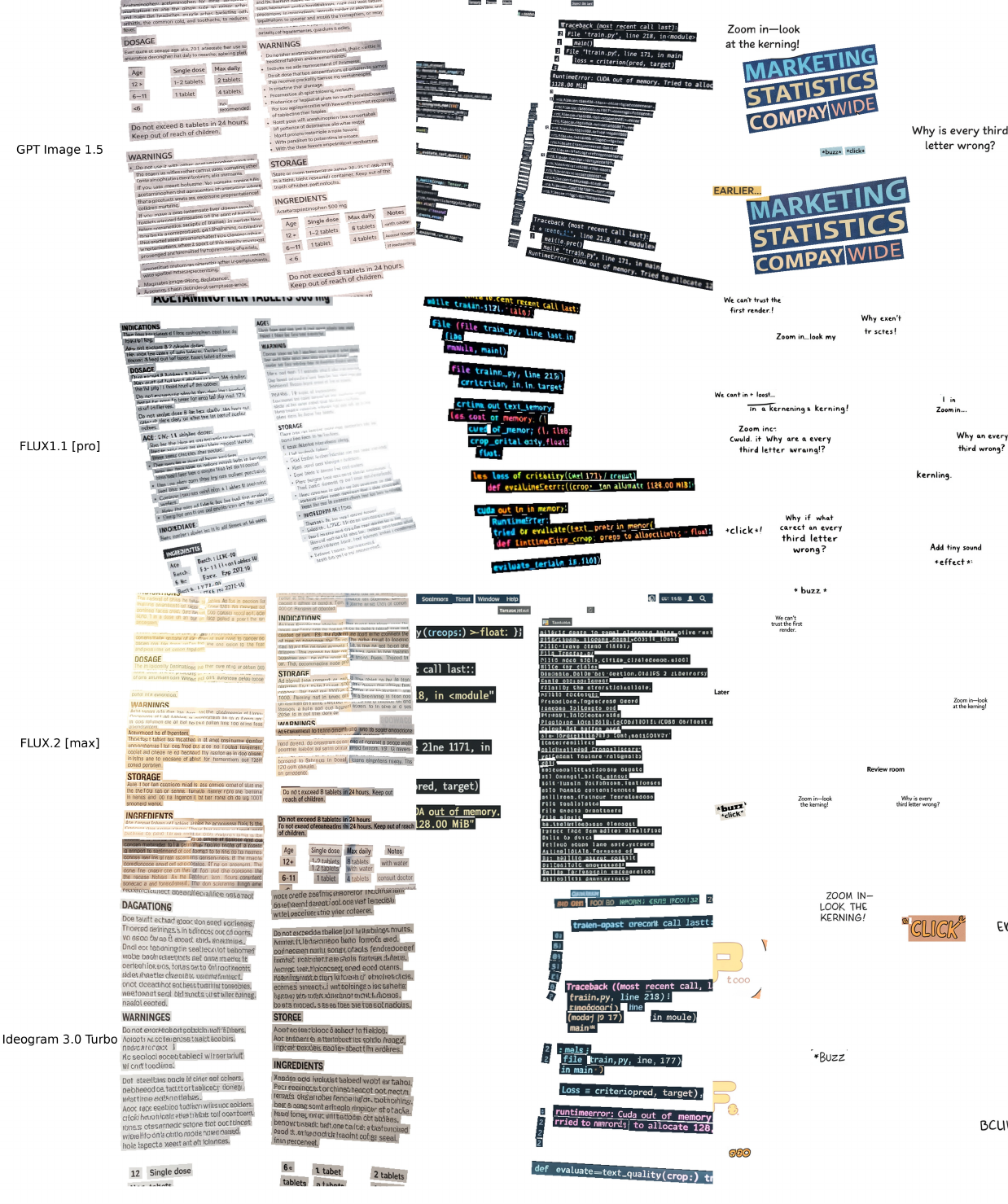} \caption{TIQA-Images examples for text quality}\label{fig:images_examples_text}
\Description{TODO}
\end{figure*}

Examples of images for the TIQA-Images dataset are presented in Figure~\ref{fig:images_examples} (the whole generated images) and Figure~\ref{fig:images_examples_text} (text-only variants). For the latter, we detected text-regions and filled in the rest of the frame with plain white. TIQA-Crops examples in Figure~\ref{fig:random_grid}

\section{Human Study Protocol}
\label{app:human}

\subsection{Participans' ability to separately visual quality from semantics}
\label{app:human:semantic}

To isolate semantic plausibility from rendering artifacts, we generate text crops with an identical prompt template, layout, and typography, varying only the target string: a real word (\texttt{world}), an anagram with identical characters (\texttt{wrodl}), and a random nonword of the same length (\texttt{wuzxh}). We use 5 text-to-image models and generate 5 images per model for each prompt, yielding diverse renderings. Then, we collect 0--5 MOS for visual text quality using the exact same protocol as TIQA-Crops (same instructions, exam, etc.).

\begin{figure}[h!]
    \centering
    \includegraphics[width=1.0\linewidth]{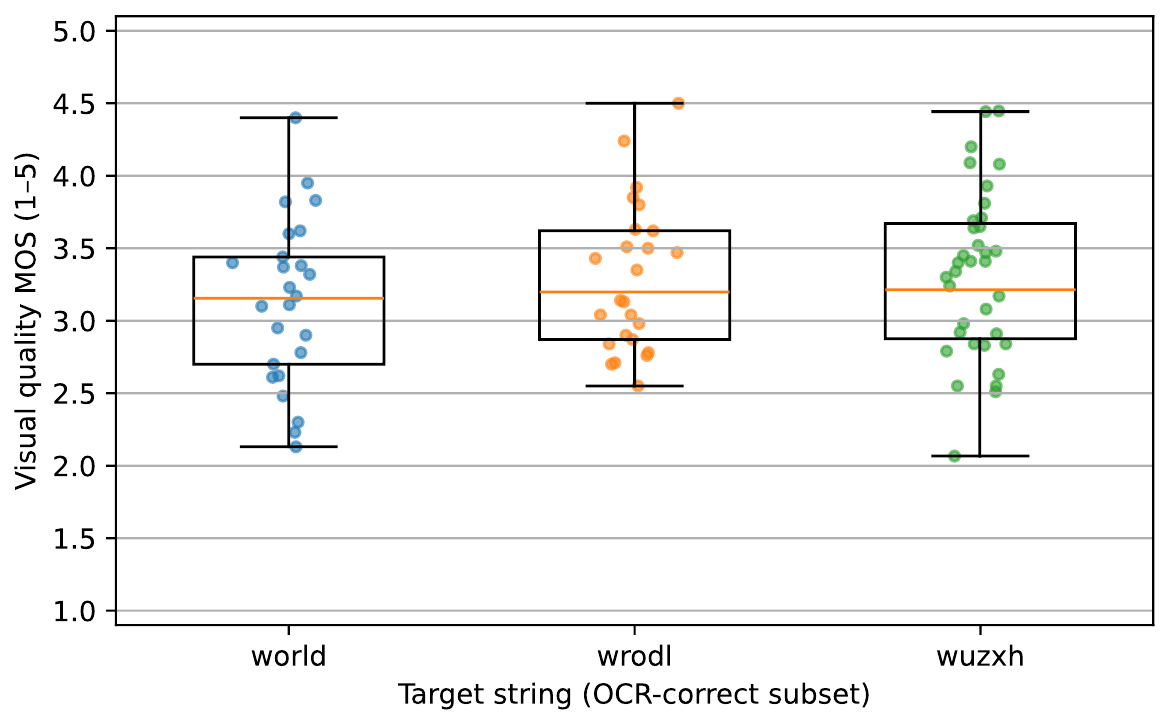}
    \caption{MOS distributions (1–5) for three target strings on the OCR-correct subset (exact transcript match). Similar distributions across a real word, an anagram, and a nonword indicate limited sensitivity of visual-quality ratings to lexical plausibility when rendering is correct.} 
    \label{fig:separ_exp}
\end{figure}

We focus on an OCR-correct subset to control for fatal rendering errors: a crop is OCR-correct if an external recognizer returns exactly the target string. On this subset, we compare MOS distributions across strings. 
Figure~\ref{fig:separ_exp} shows the MOS distributions for each prompt variant. 
The distributions largely overlap, with similar means and ranges, suggesting that lexical plausibility has limited effect on the visual text-quality MOS when using this subjective study protocol.

\subsection{Qualification Exam and Quality Control}
\label{app:human:qc}
Before starting the markup, Yandex.Tasks platform~\cite{yandex_tasks} users had to pass an exam that tested their understanding of the instructions. For both TIQA-Crops and TIQA-Images exam there were 10 demonstration crops/images, of which at least 8 had to be marked correctly. 

The markup was carried out in batches, and for both datasets a user’s answers within a batch were accepted only if they correctly answered one of the quality-control questions that were mixed into each batch.

\subsection{Annotation Statistics and MOS Computation}
\label{app:human:stats}
\begin{figure}[h!]
    \centering
    \includegraphics[width=1.0\linewidth]{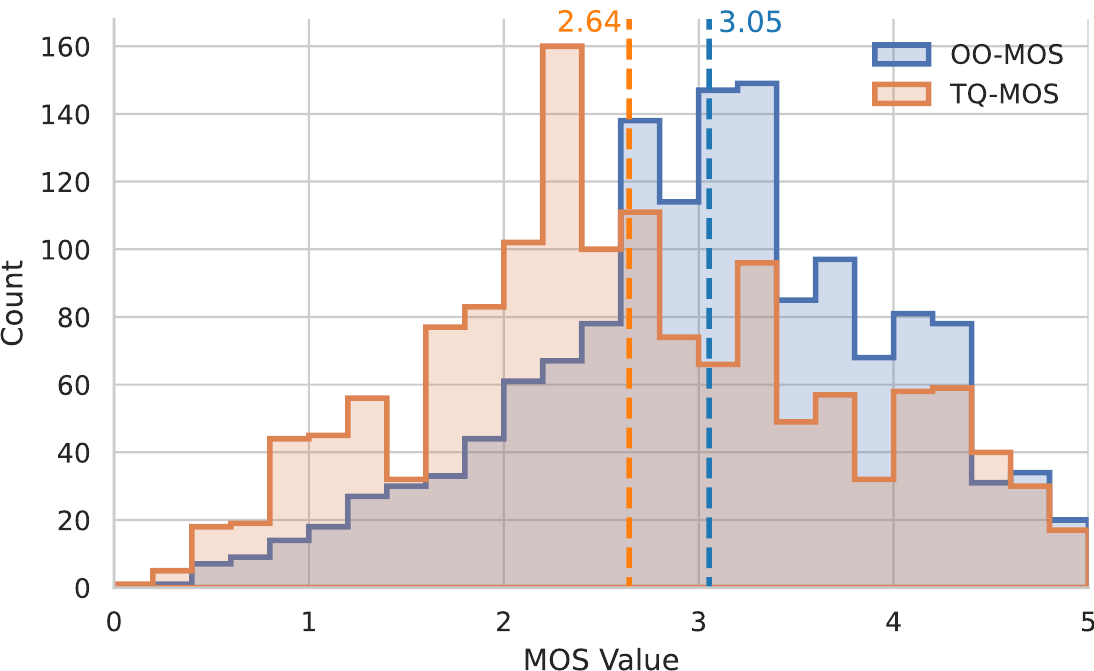}
    \caption{Distibution plot for OQ-MOS and TQ-MOS for TIQA-Images dataset. Vertical dashed lines denote mean values across all images.} 
    \label{fig:mos_dist}
\end{figure}

Each image was independently rated by 50 human raters. We report Mean Opinion Score (MOS) as a $10\%$ trimmed mean: for each image, we discard the lowest $5\%$ and highest $5\%$ of ratings (i.e., the 5 lowest and 5 highest out of 50) and average the remaining 40 ratings.
Figure~\ref{fig:mos_dist} shows distributions of MOS values for TIQA-Images dataset

\subsection{Rater Instructions}
\label{app:human:ui}

\begin{figure}[h!]
    \centering
    \includegraphics[width=\linewidth]{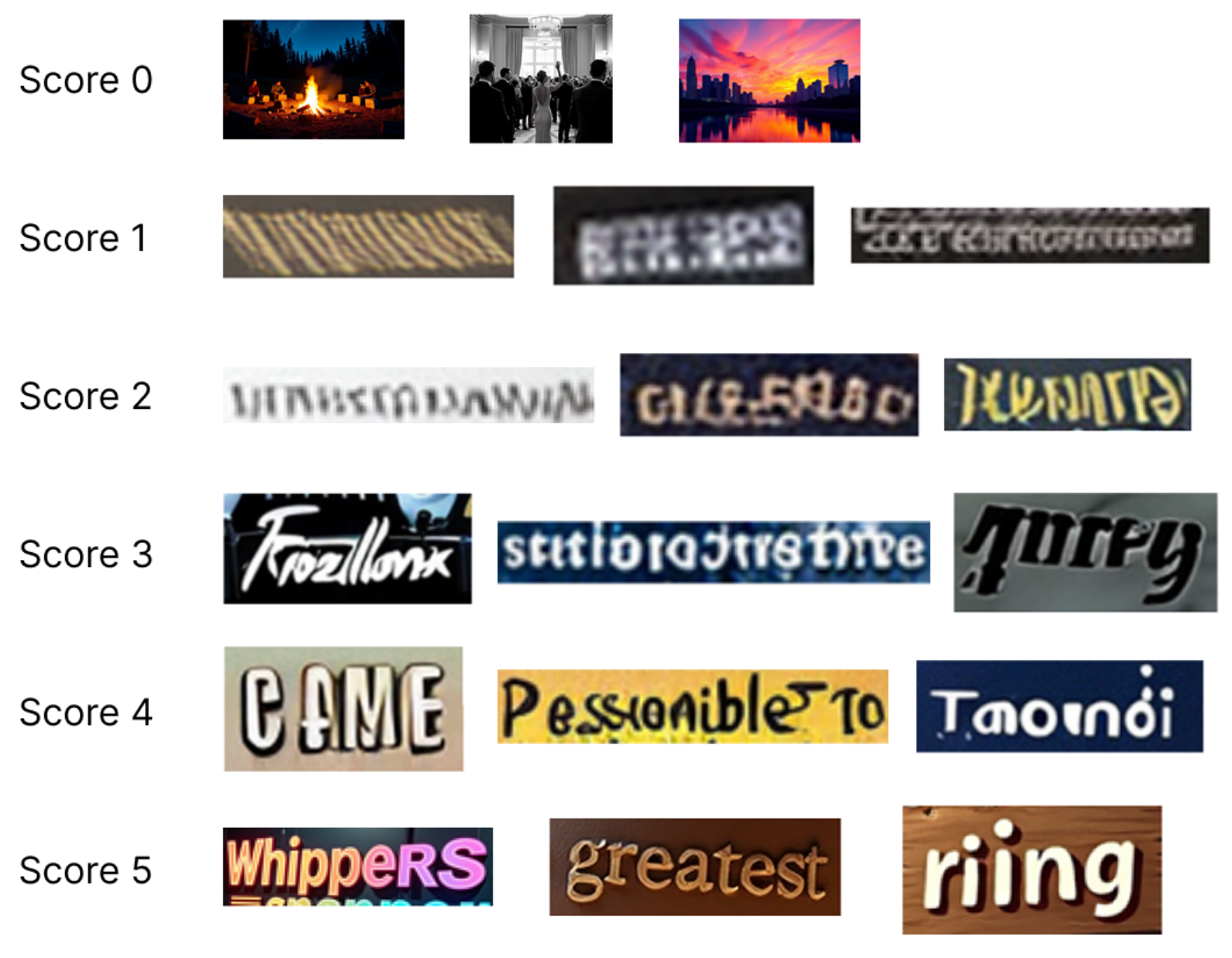}
    \caption{Representative visual examples of the rating categories shown to annotators during the TIQA-Crops labeling task. Each row corresponds to a different score on the $0$--$5$ scale and serves as a reference anchor for the rater.}
    \label{fig:crop_regression}
\end{figure}

Annotators recruited through the Yandex.Tasks platform~\cite{yandex_tasks} were provided with detailed written instructions describing the annotation protocol for both the TIQA-Crops and TIQA-Images datasets prior to participating in the study. To minimize ambiguity in the interpretation of the $0$--$5$ rating scale, each instruction set was accompanied by visual examples illustrating representative samples for every score category, ensuring that raters could anchor their judgments to concrete reference points rather than relying solely on verbal definitions. A subset of these reference examples for the TIQA-Crops rubric is shown in Figure~\ref{fig:crop_regression}. The complete instruction texts presented to the annotators are reproduced below.

\begin{figure*}[h!]
\begin{tcolorbox}[promptbox]
\textbf{Markup instructions for TIQA-Crops dataset:} 
\texttt{"Your task is to evaluate the quality of the text in the picture from 0 to 5. The images are derived from generative models, which is why there are artifacts and distortions that need to be taken into account. \\ Score 0: There is no hint of any text in the picture. \\ Score 1: The outline of the text is roughly visible, but it is completely unreadable, it is impossible to identify individual letters. The key difference from a score of 0 is that there is a general outline of the text. \\  Score 2: The text is still unreadable, but the key difference from score 1 is that individual letters begin to appear. \\  Score 3: A lot of artifacts: non-existent, stuck, floated or distorted symbols. The key difference from score 2 is that some Latin characters or similar ones are visible. Moreover, these symbols are clearer and more distinguishable. \\  Score 4: ALMOST ALL the text is perfect and readable, but there are small artifacts. In the illustration, the artifacts are highlighted in red frames. The key difference from score 3 is a lot of readable Latin letters and fewer artifacts. \\ Score 5: All the text in the picture is READABLE, each letter is in the English alphabet and is clearly visible. A low resolution is acceptable. The key difference from score 4 is the absence of obvious distortions and artifacts. \\  \\ - There are test tasks, if they are answered incorrectly, the work will not be counted. \\  - Ignore the meaning and semantics of the text. \\- Ignore the resolution of the image. That is, low resolution does not mean poor quality. The purpose of markup is to identify the presence of artifacts, distortions, and confusion. \\- Ignore the cropped pieces of other text outside the target label. \\- You can select the answer option using the keys "0", "1", etc."}
\end{tcolorbox}
\end{figure*}

\begin{figure*}[h!]
\begin{tcolorbox}[promptbox]
\textbf{Markup instructions for TIQA-Images dataset, overall quality:} \texttt{"You will see images generated by artificial intelligence, which often look high-quality, but may contain unnoticeable artifacts (including in the text). Your task is to evaluate the overall perceived quality of the entire image. \\  \\ What needs to be considered: \\ - realistic and natural images \\ - sharpness, noise, and other processing artifacts \\ - visible text artifacts are also taken into account because they affect the overall impression. \\ What NOT to consider: \\ - the image content itself \\ - the meaning of the written text \\  \\ Score 0: Very poor quality (it is difficult or impossible to make out what is shown in the image). \\  Score 1: Strong artifacts throughout the image, lots of interference. \\ Score 2: Artifacts are very noticeable (in the main text), but the image looks good enough. \\ Score 3: Some artifacts are noticeable on closer inspection. \\ Score 4: Almost the entire picture looks perfect, there are small artifacts that do not affect the overall quality. \\ Score 5: The image does not contain any artifacts, the entire text is perfectly readable. \\  \\ - There are test tasks for which the work will not be counted if answered incorrectly.  \\  \\ - Ignore the meaning and semantics of the text \\  \\ - You can select the answer option using the keys "0", "1", etc."}
\end{tcolorbox}
\end{figure*}

\begin{figure*}[h!]
\begin{tcolorbox}[promptbox]
\textbf{Markup instructions for TIQA-Images dataset, text quality:} \texttt{"You will see images that contain only text areas from the original image. Everything else has been removed and replaced with white. This is intentional. \\  \\ Your task is to evaluate the visual quality of the text itself, focusing only on the artifacts. This task does NOT concern the content of the text. \\  \\ Evaluate how visually correct and natural the text looks.: letter shape (absence of "pseudo-letters", deformed glyphs), the integrity of the strokes, interval/kerning sequence, unnatural deformations, blurring, variable thickness \\  \\ What NOT to consider \\  \\ - semantics: spelling, grammar, meaning, "does this sentence make sense?", language \\  \\ - font or styling, if they do not lead to visible distortion \\  \\ - large white areas (they are expected) \\  \\ - the overall "composition of the image" (it does not exist by design). \\  \\ How to view \\  \\ First, evaluate the image at a normal scale, then zoom in to view the smaller text. \\  \\ If the text is very small, evaluate whether it looks plausible when zoomed in. \\ Score 0: very poor quality (it is difficult or impossible to make out what is written on the image) \\ Score 1: Strong artifacts throughout the image, lots of interference and artifacts \\ Score 2: The artifacts are highly visible, but the basic inscriptions look good \\ Score 3: Some artifacts are noticeable on closer inspection, there is a lot of text without artifacts \\ Score 4: Almost all the text looks perfect, there are small artifacts that do not affect the overall quality \\ Score 5: The picture does not contain any artifacts, the entire text is perfectly readable \\  \\ - There are test tasks, if they are answered incorrectly, the work will not be counted \\ - Ignore the meaning and semantics of the text \\ - You can select the answer option using the keys "0", "1", etc."}
\end{tcolorbox}
\end{figure*}


\clearpage


\end{document}